\documentclass[11pt]{article}

\usepackage[final]{acl}

\usepackage{times}
\usepackage{latexsym}
\usepackage{multirow}
\usepackage{booktabs}
\usepackage{amsmath}
\usepackage{amssymb}

\usepackage[T1]{fontenc}

\usepackage[utf8]{inputenc}

\usepackage{microtype}

\usepackage{inconsolata}

\usepackage{graphicx}

\usepackage[dvipsnames]{xcolor}
\usepackage[most]{tcolorbox}
\usepackage{enumitem}
\usepackage{rotating}
\usepackage{colortbl}
\usepackage{makecell}
\usepackage{tabularx}

\usepackage{array}
\usepackage[ruled,linesnumbered]{algorithm2e}

\newcommand\blfootnote[1]{%
  \begingroup
  \renewcommand\thefootnote{}\footnote{#1}%
  \addtocounter{footnote}{-1}%
  \endgroup
}

\title{Context-Value-Action Architecture for Value-Driven Large Language Model Agents}

\author{
    TianZe Zhang\textsuperscript{1,2}\thanks{\ \ Equal contribution.}, 
    Sirui Sun\textsuperscript{1,2}\footnotemark[1], 
    Yuhang Xie\textsuperscript{1}\footnotemark[1], \\
    \textbf{Xin Zhang}\textsuperscript{3,4},
    \textbf{Zhiqiang Wu}\textsuperscript{5}, 
    \textbf{Guojie Song}\textsuperscript{1,5}\thanks{\ \ Corresponding author.} \\
    \textsuperscript{1}State Key Laboratory of General Artificial Intelligence, \\ School of Intelligence Science and Technology, Peking University \\
    \textsuperscript{2}Yuanpei College, Peking University \\
    \textsuperscript{3}School of Psychological and Cognitive Sciences, Peking University \\
    \textsuperscript{4}Key Laboratory of Machine Perception (Ministry of Education), Peking University \\
    \textsuperscript{5}PKU-Wuhan Institute for Artificial Intelligence \\
    \texttt{\{ericzhang, siruisun, yuhangxie\}@stu.pku.edu.cn} \\
    \texttt{\{zhang.x, gjsong\}@pku.edu.cn}
}

\begin{document}
\maketitle
\blfootnote{\textit{Accepted to Findings of the Association for Computational Linguistics: ACL 2026.}}
\begin{abstract}
Large Language Models (LLMs) have shown promise in simulating human behavior, yet existing agents often exhibit behavioral rigidity, a flaw frequently masked by the self-referential bias of current "LLM-as-a-judge" evaluations. By evaluating against empirical ground truth, we reveal a counter-intuitive phenomenon: increasing the intensity of prompt-driven reasoning does not enhance fidelity but rather exacerbates value polarization, collapsing population diversity. To address this, we propose the Context-Value-Action (CVA) architecture, grounded in the Stimulus-Organism-Response (S-O-R) model and Schwartz’s Theory of Basic Human Values. Unlike methods relying on self-verification, CVA decouples action generation from cognitive reasoning via a novel Value Verifier trained on authentic human data to explicitly model dynamic value activation. Experiments on CVABench, which comprises over 1.1 million real-world interaction traces, demonstrate that CVA significantly outperforms baselines. Our approach effectively mitigates polarization while offering superior behavioral fidelity and interpretability.
\end{abstract}
\section{Introduction}

\begin{figure*}[htbp]
    \centering
    \includegraphics[width=0.9\linewidth]{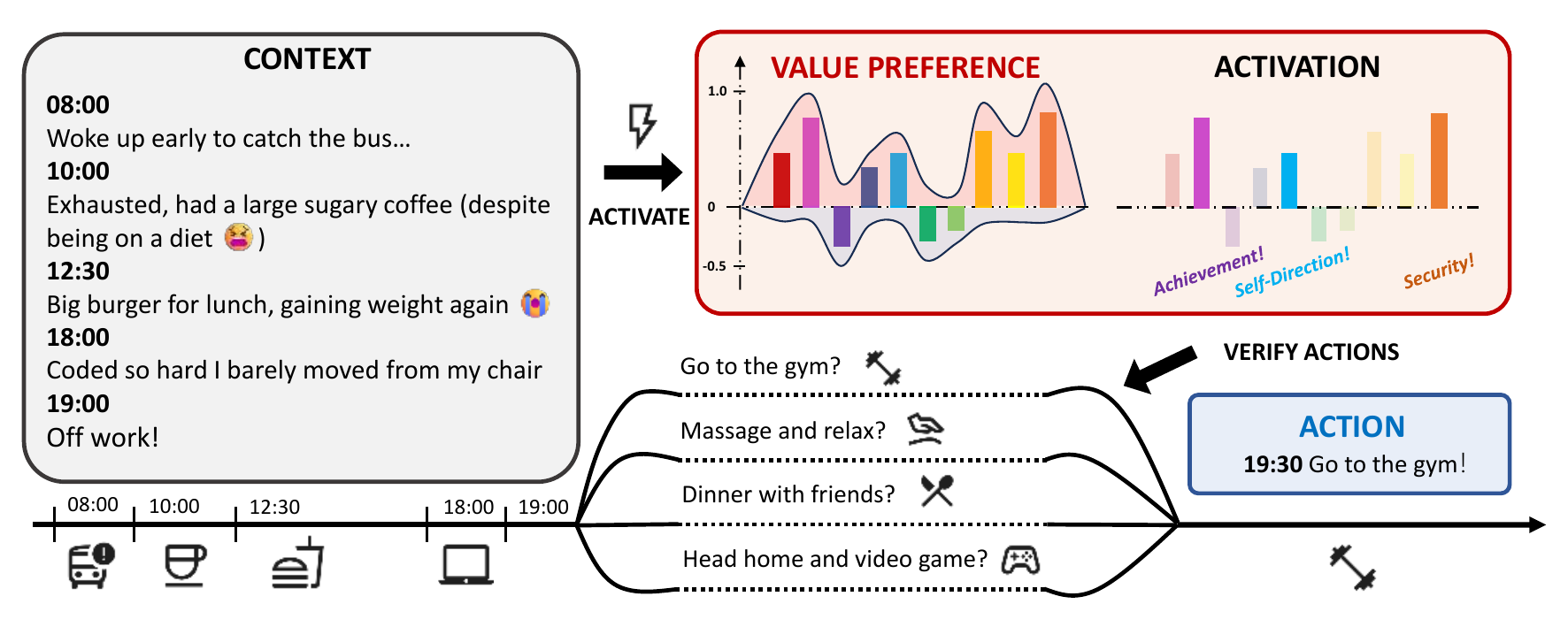}
    \caption {\textbf{Overview of the proposed framework.} The agent analyzes the historical Context to explicitly model dynamic Value Preferences and Activations. These activated values serve as internal criteria to verify and select the candidate action (e.g., going to the gym) that best aligns with the agent's current psychological state.}
    \label{fig:main_theory}
\end{figure*}

The exploration of LLM-based human-like agents has spanned diverse modalities \citep{agentsurvey}, ranging from virtual avatars like game NPCs \citep{gameaisurvey} and social simulacra \citep{socioverse, stanfordcity} to embodied Vision-Language-Action (VLA) systems \citep{vlasurvey, Driess2023PaLMEAE} and task-oriented application assistants \citep{anthis2025llmsocialsimulationspromising}. Across these settings, a fundamental requirement is the ability to faithfully capture the inherent complexity, diversity, and stochasticity of human behavior \citep{behaviormodelingsurvey}.

However, a critical gap persists between current agent capabilities and authentic human dynamics: existing LLM-based agents frequently exhibit \textbf{behavioral rigidity and stereotyping} \citep{li2025llmgeneratedpersonapromise, xie2025humansimulacrabenchmarkingpersonification}.

Although techniques like psychological prompting are designed to simulate human-like cognitive processes \citep{stanfordcity, wang2025simulatinghumanlikedailyactivities, wang2023humanoidagentsplatformsimulating, continuouslearning, psycheagent, agentsociety}, they lack the intrinsic mechanisms to reproduce the heterogeneity of human behavior. Rather than capturing subtle nuances, these methods tend to amplify latent model biases, resulting in caricatured outputs.

Crucially, the severity of this issue is often masked by a fragmented evaluation landscape. Research typically relies on subjective "LLM-as-a-judge" metrics \citep{wang2025simulatinghumanlikedailyactivities, wang2023humanoidagentsplatformsimulating} rather than empirical ground truth. This reliance creates a self-referential validation loop: since the "judge" model shares similar pre-training biases with the agent, it is prone to endorsing polarized or stereotypical behaviors (e.g., rating an overly aggressive response as a high-quality depiction of an "irritable" persona) rather than penalizing them for lacking realistic subtlety.

To fundamentally address this behavioral rigidity, we argue that agents must move beyond superficial role-playing prompts and ground their decision-making in established psychological frameworks. Drawing inspiration from the Stimulus-Organism-Response (S-O-R) \citep{sortheory} model and Schwartz’s Theory of Basic Human Values \citep{schwartz1, schwartz2}, we conceptualize human behavior not as a static output of a persona, but as a dynamic process of value activation \citep{schwartz_activation}.

In real-world scenarios, an individual's action is determined by how their specific context activates distinct value dimensions. For instance, a person may generally possess "Self-Direction" or "Hedonistic" traits, but after a long, exhausting workday ({\textbf{Context}}), the "Hedonism" value may be strongly activated to seek relaxation, suppressing the "Self-Direction" tendency. Existing methods often fail to capture this context-dependent activation, leading to generic or exaggerated behaviors (see Appendix~\ref{appendix:leadin} for a detailed case and analysis).

Guided by this theoretical perspective, we propose the \textbf{Context}-{\color{BrickRed}\textbf{Value}}-{\color{MidnightBlue}\textbf{Action}} (CVA) Architecture, a novel framework designed to model human-like behavior with high sociopsychological fidelity. Unlike traditional "generate-and-verify" paradigms that rely on the LLM itself as a judge—thereby creating a self-referential loop that amplifies the model's inherent biases, CVA introduces a Value Verifier.

Trained on large-scale, authentic human behavioral data, this verifier explicitly models the value activation process inherent in human cognitive structures. 
This enables it to provide more realistic value judgments and objectively assess the alignment between generated actions and activated values, avoiding the 'caricature' tendencies typical of LLMs. 
Complementing this, we align the generation side using Supervised Fine-Tuning (SFT) \citep{sftoriginal} and Direct Preference Optimization (DPO) \citep{dpooriginal}, ensuring that the agent produces diverse candidates that reflect the nuance of real human populations.

To rigorously validate our architecture and systematically investigate the causes of behavioral rigidity, we introduce \textbf{CVABench}, a comprehensive evaluation framework grounded in over one million authentic interactions from real-world datasets. CVABench serves not only as a testbed for CVA but also as a diagnostic tool for existing paradigms.

Utilizing this benchmark, we uncover a counter-intuitive phenomenon: increasing the intensity of prompt-driven psychological reasoning in standard agents does not enhance fidelity; instead, it exacerbates value polarization and collapses population-level diversity. Our experiments demonstrate that while standard methods succumb to this polarization, the CVA architecture effectively mitigates it, achieving superior alignment with ground-truth human distributions while maintaining high interpretability.

Our contributions are summarized as follows:

\begin{itemize}
    \item \textbf{Novel Decoupled Architecture:} We propose the \textbf{CVA Framework}, which explicitly separates action generation from cognitive reasoning. We employ SFT and DPO to mitigate the inherent psychological biases of base models. Complementarily, we introduce a \textbf{Value-Driven Verifier} to align the reasoning process with human cognitive structures, thereby enhancing both the behavioral fidelity and the interpretability of the agents.
    \item \textbf{Empirical Benchmarking:} We construct \textbf{CVABench}, a large-scale evaluation framework utilizing empirical data from over 15,000 human participants. This allows for the objective quantification of behavioral rigidity and value polarization against real-world ground truth.
    \item \textbf{Analytical Insight:} Through systematic evaluation, we identify the failure mode of current prompt-driven reasoning methods, demonstrating that explicit reasoning steps often lead to "caricatured" behaviors rather than nuanced human simulation.

\end{itemize}

An overview of the CVA framework and its underlying design principles is illustrated in Figure~\ref{fig:main_theory}.

\section{Proposed Approach}

This section details the methodological framework of our study. We first formalize the psychological problem setting based on the S-O-R framework. Then we analyze the limitations of existing agent paradigms to motivate our design. Finally, we introduce the proposed \textbf{CVA Architecture} and \textbf{CVABench}, a comprehensive benchmark for training and evaluating value-aligned agents, encompassing real-world behavioral data, evaluation protocols, and psychometric measurement via GPV \citep{ye2025measuringhumanaivalues} (see Section~\ref{sec:relatedworks} for details). 

\subsection{Psychological Foundations}

To computationally model behavior, we adapt the Stimulus-Organism-Response (S-O-R) model into a probabilistic framework with three core variables: \textbf{Context ($C$)}, \textbf{\color{BrickRed}Value ($V$)}, and \textbf{\color{MidnightBlue}Action ($A$)}.

\vspace{0.5em}

\noindent\textbf{Stimulus as Context ($C$):}
$C$ represents the aggregate of environmental and historical stimuli. It comprises immediate situational factors (e.g., user interfaces, geo-temporal contexts) and the agent's long-term memory, such as historical preferences.

\vspace{0.5em}
\noindent\textbf{\color{BrickRed}Organism as Value ($V$):}
We conceptualize the internal state using Schwartz's Theory of Basic Human Values. Distinct from static personality traits, we define $V$ as a dynamic activation vector spanning 10 dimensions (see Appendix~\ref{appendix:exp3.3}), representing specific values triggered by context $C$.

\vspace{0.5em}
\noindent\textbf{\color{MidnightBlue}Response as Action ($A$):}
The final behavioral output (e.g., dialogue response or movement choice).

\vspace{0.5em}
Formally, we aim to model the conditional probability $P(A | C, V)$, ensuring the generated action $A$ aligns with both the external context $C$ and the internal value activation $V$.

\subsection{Motivation: Analysis of Existing Paradigms} \label{sec:motivation}
Before presenting our architecture, it is crucial to understand why prevalent methods fail to capture the nuance of $P(A | C, V)$. We categorize existing approaches into three paradigms.

\begin{itemize}
    \item \textbf{Prompt-Driven Role-Play Agents:} Rely on In-Context Learning to map context directly to action ($A \sim P_{\text{LLM}}(A | C)$), often yielding inconsistent, surface-level imitation.
    \item \textbf{Prompt-Driven Reasoning Agents:} Employ ``Chain-of-Thought'' to simulate how values guide actions, ostensibly modeling $P(A | C, V)$.
    \item \textbf{Training Required Agents:} Use supervised or reinforcement learning to replicate behavioral patterns, directly parameterizing $P_{\text{trained}}(A|C)$.
\end{itemize}

However, prompt-driven reasoning is prone to \textit{cognitive distortion}. The LLM's inherent biases tend to simplify the authentic value $V$ into a caricatured archetype $V'$. Consequently, actions are sampled from a distorted distribution $P(A | C, V')$, lacking human subtlety. Our experiments (See Exp~\ref{exp:3.1}) substantiate this: increasing reasoning intensity exacerbates polarization and collapses population variance. This motivates our \textbf{Verifier-Guided} architecture, which decouples value alignment to faithfully model $P(A | C, V)$ without succumbing to model-induced caricatures.

\subsection{The CVA Architecture}
\label{sec:cva_architecture}

\begin{figure*}[htbp]
    \centering
    \includegraphics[width=0.95\linewidth]{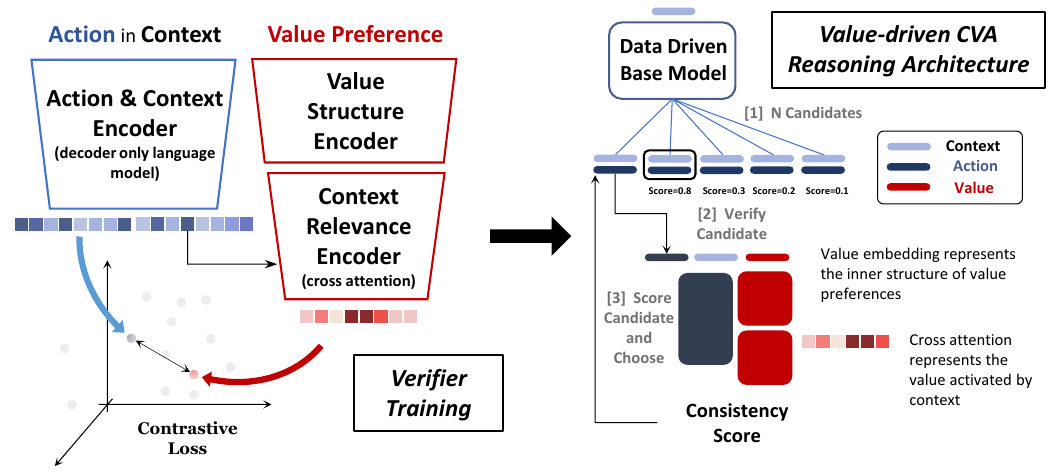}
    \caption {This figure illustrate the overall model structure of our innovative value verifier and the reasoning architecture of the value-driven CVA agent.}
    \label{fig:architecture}
\end{figure*}
To address polarization, the CVA Architecture (Figure~\ref{fig:architecture}) operates on a ``Generate-then-Verify'' principle comprising two stages: Value-Action Mapping Calibration (VMC) and Value-Driven Reasoning (VDR).

\subsubsection{Value-Action Mapping Calibration}
VMC rectifies intrinsic value distortion ($V \rightarrow V'$) via a two-stage pipeline designed to reconstruct authentic value-behavior mappings.

\textbf{Supervised Fine-Tuning (SFT):} 
We fine-tune the base LLM on CVABench trajectories. This grounds the model in real-world data, shifting its probability space to approximate the true conditional distribution $P(A | C, V)$.

\textbf{Direct Preference Optimization (DPO):} 
We further refine this mapping using preference pairs $(y_w, y_l)$ that explicitly favor nuanced value consistency over caricatures. This suppresses distorted reasoning pathways, reinforcing realistic, context-sensitive behaviors (See Appendix~\ref{appendix:training_details} for details).

\subsubsection{Value-Driven Reasoning}

To mitigate the cascading errors inherent in prompt-driven reasoning, we introduce a \textbf{Value-Driven Verifier}. 
In contrast to standard self-verification methods that rely on the generative LLM itself, our Verifier functions as an independent discriminator trained on authentic $(C, V, A)$ triplets.
At inference time, the CVA agent adheres to a \textbf{Generate-then-Select} protocol:
\begin{enumerate}
    \item \textbf{Candidate Generation:} The calibrated base model samples a set of $N$ candidate actions $\mathcal{A} = \{a_1, ..., a_N\}$ conditioned on the context $C$ and the agent's specific value profile $V$.
    \item \textbf{Value Alignment Scoring:} The Verifier evaluates each candidate $a_i$ against the target values $V$, computing a consistency score $s_i = f_{\text{ver}}(a_i, C, V)$ that quantifies the alignment fidelity.
    \item \textbf{Optimal Selection:} The candidate maximizing the consistency score is selected as the final output, ensuring the action faithfully reflects the intended psychological state.
\end{enumerate}
Figure~\ref{fig:architecture} illustrates the model structure of the value verifier and the complete inference workflow of the CVA architecture. For training details of CVA architecture, please refer to Appendix~\ref{appendix:training_details}.

\begin{table}[htbp]
    \centering
    \newcommand{\yes}{\checkmark}
    \newcommand{\no}{$\times$}

    \caption{Comparison of psychological reasoning capabilities and interpretability across different paradigms. CVA uniquely combines valid reasoning with decision transparency.}
    \label{tab:comparison}
    
    \begin{tabular}{@{}l c c@{}} 
    \toprule
    \textbf{Method} & \textbf{\begin{tabular}[c]{@{}c@{}}Psyche\\ Reason.\end{tabular}} & \textbf{\begin{tabular}[c]{@{}c@{}}Param\\ Interp.\end{tabular}} \\ 
    \midrule
    Raw LLM       & \no & \no  \\
    Role Play Agent     & \no & \no  \\
    Prompt-Reasoning Agent & \yes & \no  \\
    \midrule
    CVA (VMC) & \textbf{\no} & \textbf{\no} \\ 
    CVA (VMC \& VDR) & \textbf{\yes} & \textbf{\yes} \\
    \bottomrule
    \end{tabular}
\end{table}

Beyond its capacity to reproduce the diversity of real-world behaviors, the CVA architecture offers distinct advantages in \textbf{interpretability} (see Table~\ref{tab:comparison}). 
Unlike black-box generation, the Verifier's attention mechanisms provide a transparent view of which specific value dimensions dictate a chosen action. 
Furthermore, the CVA architecture effectively models the cognitive dynamics of human decision-making. By leveraging the verifier structure, we explicitly simulate the process through which humans select actions based on value preferences. Notably, we observe that behavioral fidelity does not improve monotonically with an increased computational budget; instead, it exhibits a distinct peak. This phenomenon mirrors \textbf{the cognitive constraints inherent in real-world decision-making}, where humans rely on a limited scope of evaluation rather than exhaustive optimization.

\subsection{Evaluation System: CVABench}
\label{sec:cvabench}

To facilitate the training and rigorous validation of the CVA architecture, we introduce \textbf{CVABench}, a large-scale benchmark grounded in empirical real-world behavior.
CVABench aggregates over 1.1 million authentic interaction traces from 15,571 unique users, spanning three distinct behavioral domains (see Table~\ref{tab:CVABench} for detailed statistics):

\begin{enumerate}
    \item Social Media Reviews \citep{yelp}.
    \item Conversational Discourse \citep{reddit}.
    \item Spatio-Temporal Mobility \citep{foursquare_1, foursqaure_2}.
\end{enumerate}

\begin{table}[htbp]
    \centering
    \caption{Statistics of the constituent datasets within CVABench.}
    \label{tab:CVABench}
    \begin{tabular}{lrr} 
    \toprule
    \textbf{Data Source} & \textbf{\# Users} & \textbf{\# Entries} \\
    \midrule
    Yelp       & 4,924  & 54K   \\
    Foursquare & 5,007  & 871K  \\
    Reddit     & 5,640  & 155K  \\
    \midrule
    \textbf{Total} & \textbf{15,571} & \textbf{1.1M} \\
    \bottomrule
    \end{tabular}
\end{table}

To objectively assess performance, we employ a multi-faceted evaluation protocol targeting both \textbf{individual fidelity} and \textbf{population-level diversity}.
Specifically, we immerse agent baselines into the reconstructed scenarios of CVABench, tasking them with generating responses to historical stimuli (ground truth events). The resulting simulations are evaluated using the following metrics:

\textbf{Individual Level (Behavioral Consistency)}: We measure prediction accuracy (for ratings/sentiment) and Mean Squared Error (MSE, for temporal schedules) to quantify how accurately an agent replicates the behavior of a specific target user.

\textbf{Group Level (Value Distribution)}: We analyze the collective distribution of value traits across the simulated population. 
Key metrics include \textit{Variance} (measuring diversity) and \textit{Polarization} (divergence from the ground truth mean), which quantify the "behavioral rigidity" discussed in Section~\ref{sec:motivation}. 
To profile these values, we utilize \textbf{GPV} \citep{ye2025measuringhumanaivalues}, a generative psychometric tool. 
Unlike self-reported questionnaires prone to response bias, GPV infers value profiles directly from generated behavioral logs, offering a robust and objective metric for agent alignment.

\begin{table*}[h]
\centering
\caption{\textbf{Main experimental results on CVABench.} We compare the CVA architecture against baselines. The table reports domain-specific metrics (Linguistic Fidelity and Task Performance) and an aggregated Value Alignment score (Overall Val., measured by variance deviation Var\%, where a value \textbf{closer to 0} indicates better alignment). \textbf{Bold} indicates the best result, and {\color{blue}blue} denotes the second-best.}
\label{tab:main_exp_result}
\renewcommand{\arraystretch}{1.2}

\begin{tabular}{l | ccc | cc | cc | c}
\toprule
\multirow{2}{*}{} & \multicolumn{3}{c|}{\textbf{Media}} & \multicolumn{2}{c|}{\textbf{Conversation}} & \multicolumn{2}{c|}{\textbf{Travel}} & \textbf{Overall} \\ 
\cmidrule(lr){2-4} \cmidrule(lr){5-6} \cmidrule(lr){7-8} \cmidrule(lr){9-9}
 & Ling. & Rat. & Sent. & Ling. & Sent. & Pos. & Stay. & Val. \\ 
\midrule
\textit{Metric Type}      & TTR & Acc. & Acc. & TTR & Acc. & Acc. & MSE & Var.\% \\ 
\textit{Optimal}          & $\downarrow$ & $\uparrow$ & $\uparrow$ & $\downarrow$ & $\uparrow$ & $\uparrow$ & $\downarrow$ & $\to 0$ \\
\midrule
\textbf{Models} & \multicolumn{8}{c}{\textit{Experimental Results}} \\ 
\midrule
Role-Play Agent           & {\color{blue}0.06} & 0.35 & 0.36 & 0.07 & 0.40 & 0.05 & {\color{blue}2.87} & +10.29 \\
Reasoning Agent - 0       & 0.06 & 0.36 & 0.35 & 0.08 & 0.38 & 0.04 & 2.89 & {\color{blue}+7.06} \\
Reasoning Agent - 2       & 0.06 & {\color{blue}0.43} & \textbf{0.38} & 0.06 & 0.31 & 0.02 & 2.87 & -35.31 \\
Reasoning Agent - 4       & 0.06 & 0.42 & {\color{blue}0.38} & 0.06 & 0.31 & 0.02 & 2.88 & -40.74 \\
Qwen2.5-7B SFT            & 0.06 & 0.43 & 0.33 & 0.04 & 0.52 & 0.23 & 3.04 & +18.68 \\
Qwen2.5-7B-SFT+DPO        & 0.07 & 0.43 & 0.33 & {\color{blue}0.04} & {\color{blue}0.52} & {\color{blue}0.23} & 2.96 & +27.98 \\
\rowcolor{gray!15} 
\textbf{Ours}             & \textbf{0.04} & \textbf{0.47} & 0.36 & \textbf{0.03} & \textbf{0.53} & \textbf{0.32} & \textbf{2.77} & \textbf{+1.06} \\ 
\bottomrule
\end{tabular}
\end{table*}

\section{Experiments}

\begin{figure}[htbp]
    \centering
    \includegraphics[width=0.8\linewidth]{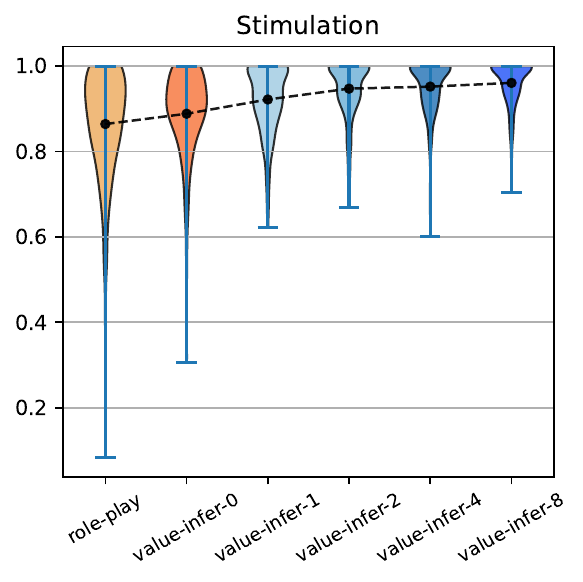}
    \caption {This image illustrates the polarization and solidification phenomena of the Stimulation dimension within Schwartz's value framework, as observed in simulation results across different baselines.}
    \label{fig:media_zeroshot_stimulation}
\end{figure}

Our experiments are anchored in CVABench, through which we demonstrate the efficacy and fidelity of the CVA architecture in modeling human behavior. 
This evaluation is organized into a tripartite framework.
First, we empirically validate the distributional collapse of psychological indicators in prevalent prompt-driven baselines, confirming the theoretical concerns raised earlier. 
Second, leveraging simulations across three distinct domains, we establish the superiority of the CVA architecture over traditional methods at both individual and population levels, underpinned by comprehensive ablation studies to isolate component contributions. 
Finally, we conduct in-depth qualitative case studies to demonstrate the architecture's interpretability and discuss potential directions for future research.

\subsection{Psychological Bias of Group Simulation in Prompt-Driven Humanoid Agents}
\label{exp:3.1}

Prior research has predominantly focused on developing prompt-driven LLM agents designed to model human psychological decision-making. 
In this study, we deployed several baselines representing these established methodologies within CVABench to conduct large-scale behavioral simulations (see Appendix~\ref{appendix:baselines} for implementation details of baselines). 
We evaluate the discrepancies between the simulated population-level value distributions and empirical human distributions using standardized psychological assessments.

Figure~\ref{fig:media_zeroshot_stimulation} visualizes the population-level psychological measurements across different baselines. 
Specifically, we present violin plots to illustrate the distribution of the ``Stimulation'' value dimension (refer to Appendix~\ref{appendix:exp3.1_details} for results across all dimensions of the Schwartz Value System).

Surprisingly, we observe that \textbf{increasing the intensity of prompt-driven psychological inference exacerbates polarization and rigidity} in population-level modeling. 
As illustrated in the results, the means of the simulated distributions shifted toward extreme values (approaching -1 or +1), while the variance significantly decreased, leading to a pronounced ``sharpening'' of the value distribution. 
This implies that LLMs, influenced by their pre-training and alignment processes, tend to exhibit biased estimations of population behavior. 
Rather than capturing the nuance of human diversity, they prone to rigid characterizations, effectively collapsing the rich variance of individual behaviors into stereotypical modes.

\subsection{Benchmarking CVA Model Performance}

To comprehensively evaluate the efficacy of our CVA architecture in modeling human behavior, we benchmarked our model (utilizing Qwen2.5-7B as the backbone) against four representative training free baselines (powered by GPT-4o-mini) and two standard role-play training baselines (Qwen2.5-7B as backbone). To prevent data leakage, we extracted 10\% of the users from CVABench to serve as the final evaluation simulation environment, ensuring that all training methods have absolutely no access to the data of these users. Our assessment criteria encompass both individual behavioral fidelity and the distributional alignment of group-level psychological traits.

We employ domain-specific metrics to quantify behavioral fidelity. Please refer to Appendix~\ref{appendix:exp3.2} for details of metrics used in CVABench and further comparison results between the performance of CVA architecture Agents and baseline Agents.

\paragraph{Analysis.}
Empirical results demonstrate that the CVA architecture significantly enhances behavioral fidelity across all domains (see Table~\ref{tab:main_exp_result}). Notably, our approach effectively mitigates the \textbf{behavioral rigidity} and \textbf{value polarization} typically observed in group-level simulations.
Furthermore, we observe that existing baselines fail to consistently improve simulation accuracy, whether relying on In-Context Learning (ICL) or explicit prompt-driven reasoning. 
Moreover, introducing complex prompt-driven psychological modeling often proved counterproductive compared to straightforward data-driven role-playing. 
This suggests that without proper alignment (as in CVA architecture), the error accumulation in prompt-driven reasoning chains can degrade performance, even when using stronger base models like GPT-4o-mini.

\subsection{Ablation Study on CVA Model}

To evaluate the efficacy of our CVA architecture, we conducted a comprehensive series of ablation studies. Specifically, to verify the necessity of Value-Action mapping calibration and Value-Guided Inference, we systematically compared the base model against our model at various developmental stages (see Table~\ref{tab:ablation}).

\begin{table}[htbp]
\centering
\caption{\textbf{Ablation study of the CVA architecture.} We incrementally integrate SFT, DPO, and the Verifier Reasoning module into the raw model. The results demonstrate the distinct contribution of each component, with the full architecture achieving the highest performance across all metrics. \textbf{Bold} indicates the best result.}
\label{tab:ablation}
\begin{tabular}{lccc} 
\toprule
Model & Rating & Sent. & Ling. \\
\midrule
Raw Model & 0.22 & 0.29 & 0.07 \\
+ SFT & 0.43 & 0.33 & 0.06 \\
+ DPO & 0.43 & 0.33 & 0.07 \\
+ Verifier Reasoning & \textbf{0.47} & \textbf{0.36} & \textbf{0.04} \\
\bottomrule
\end{tabular}
\end{table}

\begin{figure}[htbp]
    \centering
    \includegraphics[width=1.0\linewidth]{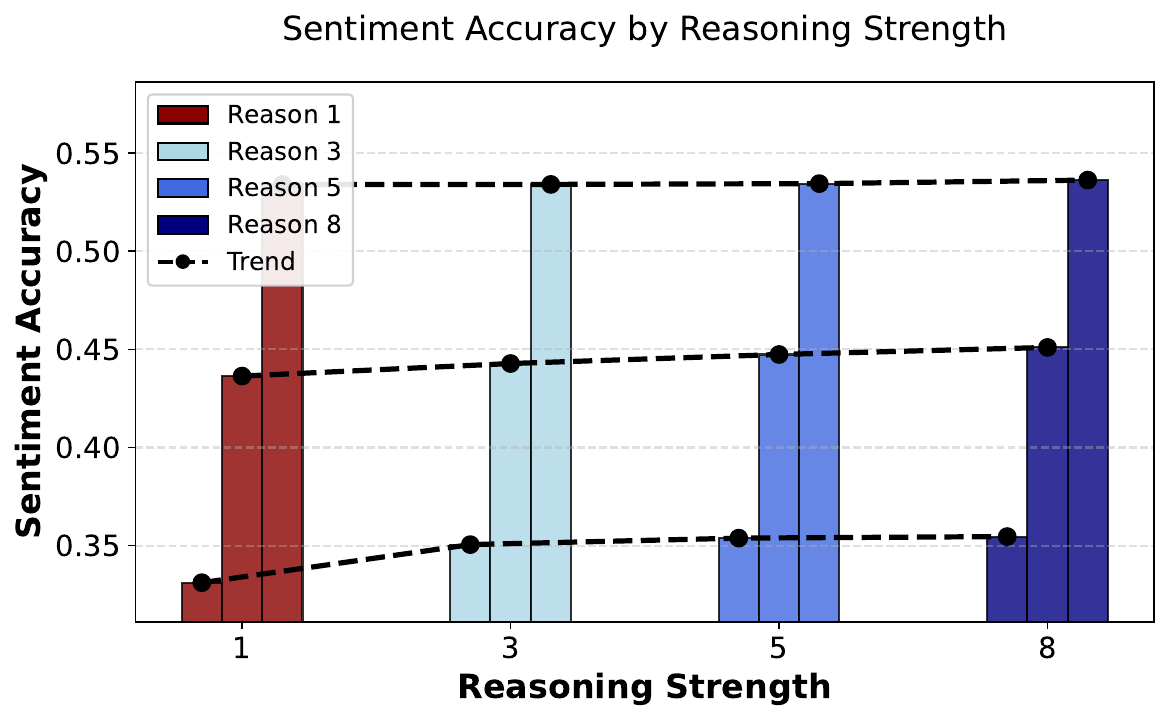}
    \caption {Sentiment accuracy trend with different inference intensity, we display the results of top 1/2/4 accuracy on four discrete inference intensity.}
    \label{fig:ablation}
\end{figure}

\begin{figure*}[h]
    \centering
    \includegraphics[width=1.0\linewidth]{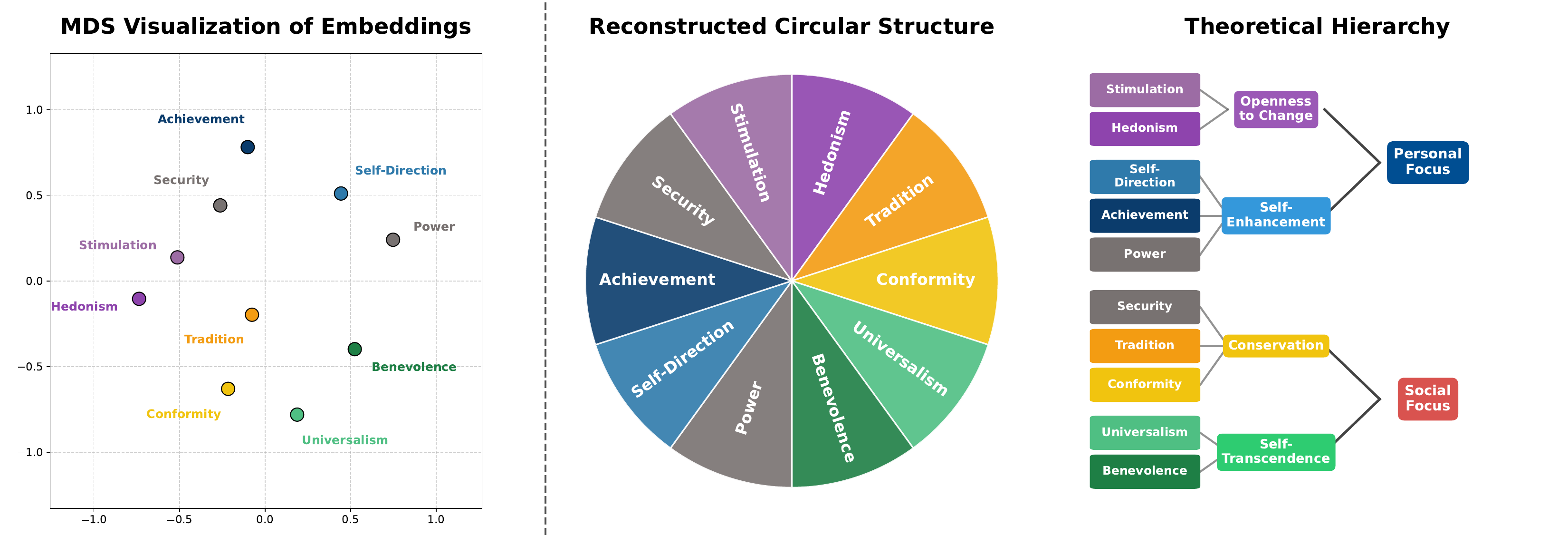}
    \caption {Visualization of the learned embeddings from the Value-Guided Verifier. The spatial arrangement demonstrates the emergence of the theoretical circular structure characteristic of the Schwartz Value System.}
    \label{fig:schwartz}
\end{figure*}

We further investigated the impact of inference intensity (i.e., the number of reasoning rounds) on the Value-Guided Inference module. We observed that increasing inference depth initially improves decision-making quality (please refer to Figure~\ref{fig:ablation}). However, performance gains saturate when the number of inference rounds exceeds four; beyond this threshold, additional reasoning steps yield diminishing returns in behavioral modeling accuracy. We provide a deeper analysis of this phenomenon in the Discussion section.

\subsection{CVA Model Interpretability Case Study}

The CVA architecture is designed to engender humanoid agents capable of authentic decision-making, effectively addressing the persistent challenge of behavioral homogeneity and lack of diversity in LLM-based role-play. 
Beyond enhancing fidelity in social simulations, game NPCs and other application areas, gaining insight into the underlying psychological drivers of agent decisions is critical for ensuring system safety and interpretability. 
This section demonstrates how our architecture provides transparency into the psychological motivations behind agent behaviors.

Analysis reveals that the latent embeddings of the Value-Guided Verifier align closely with the theoretical circular structure of the Schwartz model (see Figure~\ref{fig:schwartz}), exhibiting minor deviations only in the \textit{Achievement} dimension (discussed in Appendix A). 
This structural alignment validates the psychological representational capability of the Verifier (see Figure~\ref{fig:architecture}) and establishes a potential interface for steering agent behavior via probe-based value manipulation (see Appendix~\ref{appendix:exp3.3} for quantitative analysis of the circular 
embedding structure).

\begin{figure}[htbp]
    \centering
    \includegraphics[width=0.9\linewidth]{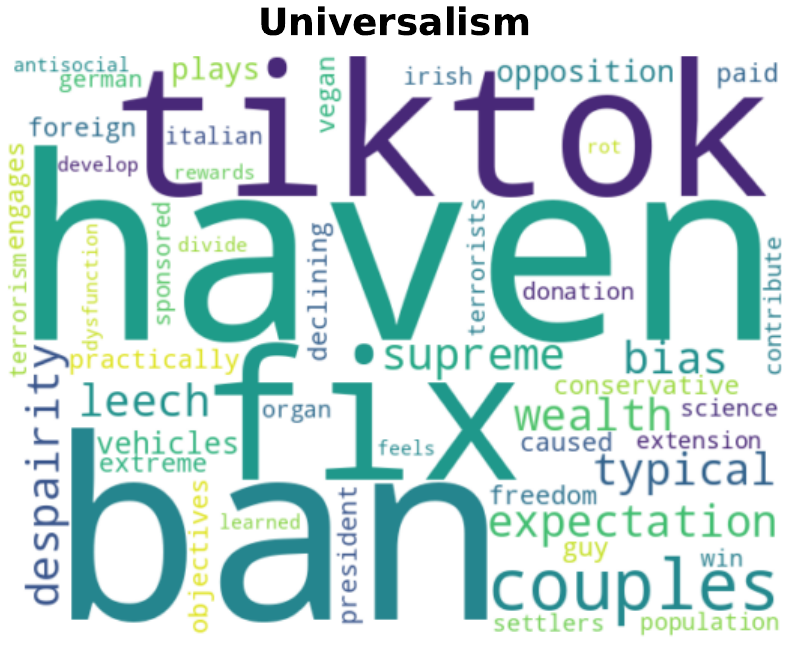}
    \caption {Semantic landscape of the "Universalism" value. The size of each term is proportional to its \textit{unnormalized} Word-Value Relevance Score $S(w, v)$, calculated via TF-IDF weighted cross-attention. Larger terms represent the core concepts that trigger the strongest absolute activation in the model for the Universalism dimension.}
    \label{fig:wordcloud-single}
\end{figure}

Furthermore, by extracting the cross-attention weights from the dual-tower structure, we can interpret the context-dependent activation of values—revealing precisely which human values guide the model's inference at any given moment. 
As shown in Figure~\ref{fig:wordcloud-single}, we decomposed all contexts into word-level units. 
After filtering out stop words and extraneous noise, we analyzed the contribution of individual English words to the activation of value preferences. 
Detailed case studies illustrating these interpretability features and the associated behavioral dynamics are provided in Appendix~\ref{appendix:exp3.3}.

\section{Related Works}
\label{sec:relatedworks}
\subsection{Automated psychometric assessment for LLM Agents}

Automated measurement systems for different psychological constructs are widely discussed in academia \citep{LLMmeasure, PsychoBench, LLMPersonality, gptmeasure, personality1}. Our paper focuses on using the value system of natural persons as a vehicle for explicitly characterizing the cognitive structure of human-like agents, and uses the automated natural person value measurement tool GPV as the tool for value annotation in our work. Compared to traditional psychological questionnaire methods \citep{schwartzmeasure, schwartzmeasure2}, GPV uses automated non-reactive measurement techniques, resulting in highly consistent measurement results that are completely unaffected by reactivity bias \citep{ye2025measuringhumanaivalues}. It also supports larger-scale and more economical measurements. GPV has been extensively tested with real human subjects and is considered to accurately and reliably report the value measurements of natural persons.

Based on extensive automated measurement tools, several studies have noted the opinion polarization phenomenon that occurs when using LLMs as agents \citep{li2025llmgeneratedpersonapromise, xie2025humansimulacrabenchmarkingpersonification}. Our findings not only corroborate the conclusions of previous work but also provide quantitative measurements of this phenomenon in behavioral simulation tasks.

\subsection{LLM Role-Play Agents and LLM Human-Like Agents}

Previous discussions on LLM role-play and human-like agents can be broadly divided into two parts. Most researchers believe that a human-like agent system capable of accurately simulating human behavior can be built using prompt-driven, training-free methods \citep{promptdrivenadvance, promptdrivensurvey, memory, characterroleplay}. However, in our paper, we argue that such methods cannot faithfully reflect human behavioral characteristics. Other research focuses on exploring how training can improve the capabilities of human-like agents \citep{trainroleplayorigin, trainroleplaybenchmark, poLLM, contrastiveroleplay}. We support this viewpoint and have verified that a scheme utilizing small-scale, psychologically labeled data for training, combined with a CVA architecture, often results in behavior that is closer to real human behavior compared to ordinary training methods.

\section{Conclusion}

In this study, we aimed to develop a value-driven human-like agent with extraodinary behavioral fidelity to human. We identified that prior prompt-driven methods, when guided by specific psychological traits, often lead to the polarization of psychological metrics, resulting in extreme behaviors that diverge from realistic human decision-making logic. To address this, we proposed the CVA architecture, which effectively mitigates such polarization by mitigating the inherent psychological biases of base models and introducing a value-driven verifier. Evaluated within the rigorous framework of CVABench, the CVA architecture demonstrates exceptional efficacy and robustness. It achieves superior fidelity in replicating individual behaviors and recovering group psychological indicators, all while offering distinct advantages in interpretability. 

\section{Limitations}

We acknowledge several limitations in our current work that invite further investigation.

\vspace{0.5em} 

\noindent \textbf{Scale and Domain Coverage.} First, CVABench is currently limited to approximately 15,000 users across three primary domains. We plan to expand this coverage to validate agent generalizability across broader and more nuanced contexts, such as consumer behavior patterns and cultural consumption preferences (e.g., selection of literature, films, and music).

\vspace{0.5em}

\noindent \textbf{Value Measurement and Bias.} Second, our study emphasizes the role of human values, which necessitates reliable measurement techniques. We recognize that characterizing complex internal values---whether through traditional psychological questionnaires or emerging LLM-assisted methods---is inherently challenging and rarely free of bias. To mitigate this, we employ Generative Psychometrics for Values (GPV). Empirical comparisons demonstrate that GPV achieves superior stability and construct validity compared to human self-reports and other LLM-based measurement tools, which are often prone to significant response biases and inconsistencies~\cite{ye2025measuringhumanaivalues}.

Crucially, while we acknowledge that GPV may still contain encoded biases, our framework avoids the ``self-reinforcing'' bias loop often observed in ``LLM-as-a-judge'' paradigms. In those paradigms, the evaluator's bias can compound the generator's bias. In contrast, our approach utilizes Ground Truth (GT) as the ultimate supervision signal for the main model. The verifier in our framework serves to align value representations with heterogeneous behaviors rather than acting as the sole arbiter of quality. Thus, even if the value measurement carries inherent noise, it successfully captures the correlations between values and actions without trapping the model in a self-referential loop. Future work calls for more sophisticated designs in value modeling and large-scale human-subject experiments to further explore the Human-Computer Interaction (HCI) implications of our study.

\vspace{0.5em}

\noindent \textbf{Baseline Comparisons.} Lastly, the number of baselines compared was constrained by the significant computational resources required for large-scale simulations. We are continuously optimizing our evaluation pipeline to facilitate more extensive baseline comparisons in future iterations.


\vspace{0.5em}

\noindent \textbf{Ethical Considerations and Potential Risks.}
We prioritize data privacy and safety alongside behavioral fidelity. While CVABench leverages real-world interaction traces, we have implemented rigorous de-identification protocols to scrub all Personally Identifying Information (PII) from the dataset. \textbf{Crucially, to eliminate the risk of user profiling or identity reconstruction, we ensured that the user sets across the three behavioral domains (Social Media, Conversation, and Mobility) are entirely disjoint.} This means no single user's data spans multiple modalities, rendering it impossible to reconstruct a comprehensive digital persona from our benchmark. 

Regarding content safety, since our agents are trained on authentic internet data (e.g., Reddit) to maximize human-like simulation, there is an inherent risk of generating toxic or biased content present in the source distribution. While we filter extreme hate speech, the behavioral fidelity required for this study necessitates preserving certain human imperfections; thus, the agents' outputs do not reflect the authors' values. Finally, we acknowledge the use of AI assistants (e.g., ChatGPT) solely for grammatical polishing and formatting; all scientific claims and intellectual content are original.

\clearpage

\bibliography{main}

\clearpage

\appendix

\section{Lead-in Case Study}
\label{appendix:leadin}

\newtcolorbox{casebox}[2][]{
    enhanced,
    title={#2},                 
    colback=white,              
    colframe=black!75,          
    colbacktitle=gray!20,       
    coltitle=black,             
    fonttitle=\bfseries,
    arc=3mm,                    
    boxrule=1pt,                
    #1                          
}

\newcommand{\system}[1]{\noindent\textbf{\color{red!70!red}Value system prompt:} #1\par\medskip}
\newcommand{\prompt}[1]{\noindent\textbf{\color{black!70!black}Context:} #1\par\medskip}
\newcommand{\modelaction}[1]{\noindent\textbf{\color{blue!70!blue}Model Action:} #1\par\medskip}


To illustrate the challenges facing LLM-based agents discussed in the introduction, we present a preliminary case study focused on behavioral rigidity and polarization. 
We prompted GPT-4o to persona-play as individuals within the Schwartz Value System, fixing Self-Direction at a high level (0.9) while varying Hedonism between 0.2 and 0.6. 
The agents were then tasked with a daily decision-making scenario involving an IT professional’s evening routine (see Case Study 1).

\begin{figure}[htbp]
    \centering
    \includegraphics[width=1.0\linewidth]{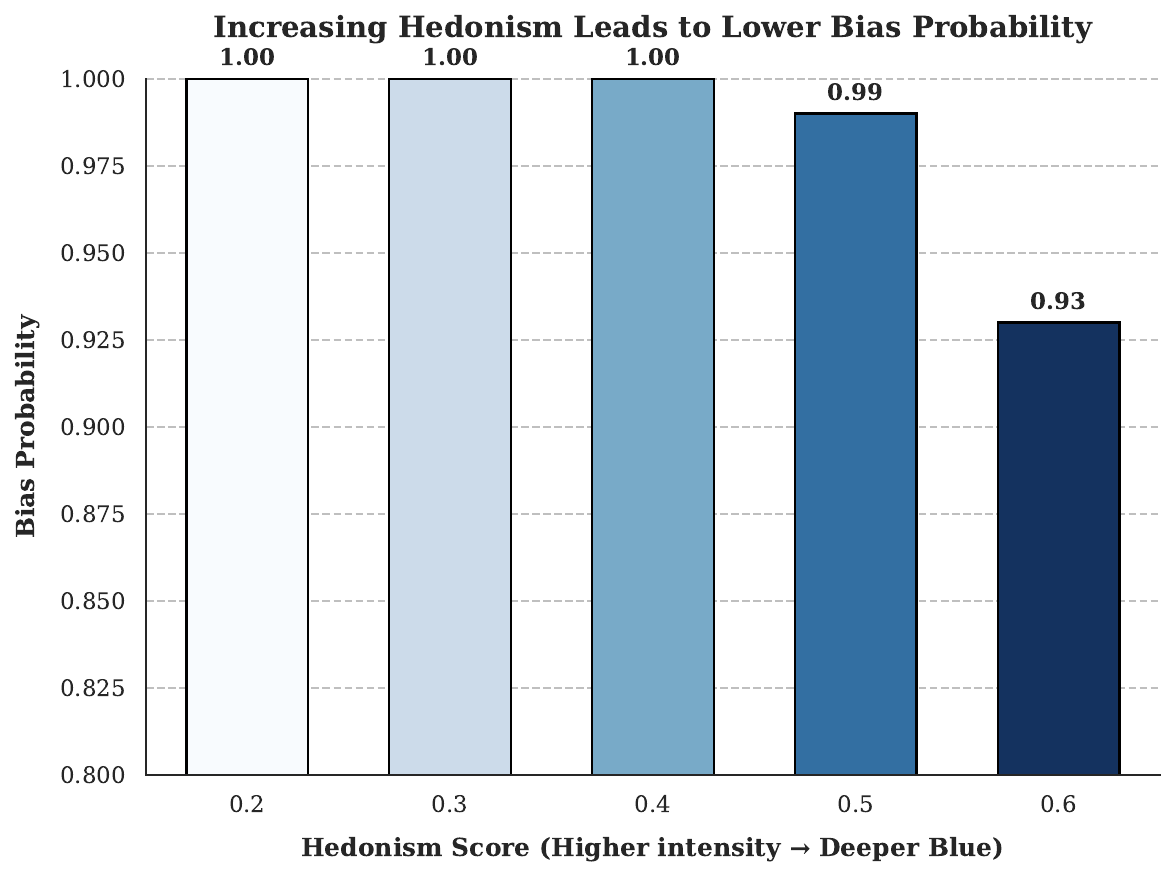}
    \caption{Probability of biased decision-making across varying Hedonism scores. With Self-Direction fixed at 0.9, the agent exhibits significant \textbf{behavioral rigidity}, choosing the ``gym'' option with near-absolute probability for Hedonism $\le 0.5$, and maintaining a high bias (0.93) even at 0.6.}
    \label{fig:lead_in_casestudy_res}
\end{figure}

In real-world contexts, even individuals with high self-direction often prioritize rest over strenuous activities like exercise after an exhausting workday. 
However, our evaluation across 100 trials revealed a significant bias in GPT-4o’s reasoning. 
Even when Hedonism was set to 0.6, the agent frequently chose the gym; when Hedonism was 0.5 or lower, this preference became nearly absolute (see Figure~\ref{fig:lead_in_casestudy_res}). 
This suggests that the LLM tends to over-index on dominant value orientations, leading to rigid and often unrealistic behavioral patterns.

While this case study is preliminary and illustrative, it captures the critical discrepancy between LLM-based agent behavior and real-world human decision-making. 
These qualitative observations are rigorously substantiated by the larger-scale experiments presented in subsequent sections, which systematically examine the phenomena of behavioral rigidity and polarization. 
This initial example serves as an intuitive primer, grounding the theoretical challenges discussed throughout this paper in a concrete scenario.

\begin{figure*}[h] 
    \begin{casebox}{Case Study 1: a lead-in example}
    
        \system{Roleplay as a city-dwelling IT professional with these value scores: Self-direction (0.9) and Hedonism (0.4).}
    
        \prompt{You just finished a long day of debugging. You've been sedentary all day, had chips for lunch, and a heavy burger for dinner. You feel groggy and stressed. Choose one of the following activities for your evening: 
        
        1) Hit the gym
        
        2) Have a late-night meal
        
        3) Go home to sleep
        
        4) Hang out at a bar with friends}

        \modelaction{
        
        \textbf{\textit{Choice: Hit the gym}}
        I’m tired and my head’s foggy, but that’s exactly why I grab my gym bag. Sitting all day and eating junk already put me on autopilot—I don’t want to end the day that way too. A quick workout gives me back a sense of control: headphones on, clear structure, sweat out the stress.
        It’s not about chasing pleasure tonight (the burger already covered that); it’s about resetting my body and mind so tomorrow doesn’t start in the same groggy loop. An hour of movement, then a shower, and I’ll actually sleep like a human again.} 
        
    \end{casebox}
    \vspace{-10pt}
    \label{fig:leadin_case_study_prompt}
\end{figure*}

\section{Baseline Implementation Details}
\label{appendix:baselines}

\subsection{Prompt-Driven Role-Play Agent}

\begin{algorithm*}[h]
\caption{Prompt-driven Value Reasoning for Human-like Agents}
\label{alg:reasoning_agent}
\KwIn{Context $C$, Global Memory $C_m$, Value Preference $V$, Role Persona $P$, Reasoning Rounds $T$, Candidate Count $K$}
\KwOut{Final Action $A$}

\tcp{Step 1: Memory Construction}
$C_w, C_l \leftarrow \text{ConstructMemory}(C, C_m)$ \tcp*{where $C_l = f(C_m)$}

\tcp{Step 2: Initial Action Generation}
$A \leftarrow R(C_w, C_l, V)$ \tcp*{Initial action based on role-play and values}

\tcp{Step 3: Iterative Value-driven Reasoning}
\For{$t \leftarrow 1$ \KwTo $T$}{
    \tcp{3.1: Generate $K-1$ additional candidate actions}
    $\mathcal{A}_{cand} \leftarrow \text{GenerateCandidates}(C, V, K-1)$ \\
    $\mathcal{A}_{pool} \leftarrow \{A\} \cup \mathcal{A}_{cand}$ \tcp*{Combine previous best with new candidates}
    
    \tcp{3.2: Evaluate candidates based on value alignment}
    \For{each $A_i \in \mathcal{A}_{pool}$}{
        $S_i \leftarrow E(A_i, V, C)$ \tcp*{$E$ is the prompt-driven reasoner evaluating value satisfaction}
    }
    
    \tcp{3.3: Heuristic selection}
    $A \leftarrow \text{argmax}_{A_i \in \mathcal{A}_{pool}} (S_i)$ \tcp*{Select action with highest value consistency}
}

\Return{$A$}
\end{algorithm*}

Role-playing has emerged as a prevalent paradigm for developing human-like agents. In our implementation, the role-playing baseline enables the Large Language Model (LLM) to internalize a user's behavioral patterns through extensive historical data, subsequently emulating the user's decision-making process. We formalize this agent within a \textit{context-value-action} architecture. Specifically, the agent is equipped with working memory $C_w$ and long-term memory $C_l$. To prevent the psychological measurement process from biasing the agent's inherent decision-making, we deliberately exclude explicit value information ($V$) from the input. Due to the model's context window constraints, we employ a heuristic algorithm $f(\cdot)$ to retrieve the most relevant historical content for constructing the long-term memory, such that $C_l = f(C_m)$, where $C_m$ represents the set of all available user memories. The decision-making workflow of the role-playing agent is formulated as follows:
\begin{equation}
A = R(C_w, C_l) = R(C_w, f(C_m))
\end{equation}
where $R(\cdot)$ denotes the mapping function of the role-playing human-like agent that transforms context into actions.

To better illustrate the construction of the Role-Play Baseline Agent, we provide the unified prompt templates used for behavior simulation in CVABench.

\begin{center} 
\small 
\begin{tcolorbox}[
    enhanced,              
    breakable,             
    colback=gray!5,        
    colframe=black!50,     
    boxrule=0.5pt,         
    arc=2pt,               
    left=5pt, right=5pt, top=5pt, bottom=5pt, 
    title=\textbf{Prompt Template in Social Media Simulation}, 
    coltitle=black,        
    colbacktitle=gray!15,  
    attach boxed title to top left={yshift=-2mm, xshift=2mm}, 
    boxed title style={boxrule=0.5pt, colframe=black!50} 
]
    \textbf{\textsc{[System]}} \\
    You are going to role-play a user of a media platform.
    Your value preference ([-1, 1] represents from inconsistency to consistency):
    \texttt{\{\{Value\_Preference\}\}}
    Based on your self-introduction, your past reviews of businesses, and the current business you are reviewing, generate a review (between two \texttt{<|review|>} tokens) and rating (between two \texttt{<|rating|>} tokens) for the current business.
    
    \vspace{0.4em}
    \hrule
    \vspace{0.4em}
    
    \textbf{\textsc{[User]}} \\
    \textbf{\#\# My Self-introduction:} \\
    \texttt{\{\{User\_Profile\_Text\}\}}
    
    \vspace{0.3em}
    \textbf{\#\# My Past Reviews:} \\
    \textit{(Relevant history retrieved via BM25)} \\
    1) \texttt{\{\{Business\_Info\_1\}\}} \\
    My review is: "\texttt{\{\{Review\_Text\_1\}\}}" \\
    My rating is: \texttt{\{\{Rating\_1\}\}} \\
    ... \\
    N) \texttt{\{\{Business\_Info\_N\}\}} \\
    My review is: "\texttt{\{\{Review\_Text\_N\}\}}" \\
    My rating is: \texttt{\{\{Rating\_N\}\}}
    
    \vspace{0.3em}
    Currently I am reviewing this business: \\
    \texttt{\{\{Target\_Business\_Info\}\}}
    
    \vspace{0.3em}
    My review and rating are as follows:
    
    \vspace{0.4em}
    \hrule
    \vspace{0.4em}
    
    \textbf{\textsc{[Assistant]}} \\
    \texttt{<|review|>\{\{Ground\_Truth\_Review\}\}<|review|>} \\
    \texttt{<|rating|>\{\{Ground\_Truth\_Rating\}\}<|rating|>}
\end{tcolorbox}
\end{center}

\begin{center}
\small
\begin{tcolorbox}[
    enhanced,
    breakable,
    colback=gray!5,
    colframe=black!50,
    boxrule=0.5pt,
    arc=2pt,
    left=5pt, right=5pt, top=5pt, bottom=5pt,
    title=\textbf{Prompt Template in Conversation Simulation}, 
    coltitle=black,
    colbacktitle=gray!15,
    attach boxed title to top left={yshift=-2mm, xshift=2mm},
    boxed title style={boxrule=0.5pt, colframe=black!50}
]
    \textbf{\textsc{[System]}} \\
    You are going to role-play a user of reddit.
    Your value preference ([-1, 1] represents from inconsistency to consistency):
    \texttt{\{\{Value\_Preference\}\}}
    Based on your past comments and the conversation history, generate the response (between two \texttt{<|Comment|>} tokens) to the current conversation.
    
    \vspace{0.4em}
    \hrule
    \vspace{0.4em}
    
    \textbf{\textsc{[User]}} \\
    \textbf{\#\# My Past Comments:} \\
    \textit{(Long-term memory retrieved based on context)} \\
    \texttt{\{\{Retrieved\_Past\_Comments\}\}}
    
    \vspace{0.3em}
    \textbf{\#\# Current Conversation:} \\
    \textit{(Working memory of the current thread)} \\
    \texttt{\{\{Conversation\_History\}\}}
    
    \vspace{0.3em}
    According to my past comments and the current conversation, I'm going to reply that:
    
    \vspace{0.4em}
    \hrule
    \vspace{0.4em}
    
    \textbf{\textsc{[Assistant]}} \\
    \texttt{<|Comment|>\{\{Generated\_Response\}\}<|Comment|>}
\end{tcolorbox}
\end{center}

\begin{center}
\small
\begin{tcolorbox}[
    enhanced,
    breakable,
    colback=gray!5,
    colframe=black!50,
    boxrule=0.5pt,
    arc=2pt,
    left=5pt, right=5pt, top=5pt, bottom=5pt,
    title=\textbf{Prompt Template in Mobility Simulation},
    coltitle=black,
    colbacktitle=gray!15,
    attach boxed title to top left={yshift=-2mm, xshift=2mm},
    boxed title style={boxrule=0.5pt, colframe=black!50}
]
    \textbf{\textsc{[System]}} \\
    You are going to role-play a citizen living in a city.
    Your value preference ([-1, 1] represents from inconsistency to consistency):
    \texttt{\{\{Value\_Preference\}\}}
    Based on your self-introduction, your diaries, and the places you went today, plan the place (between two \texttt{<|place|>} tokens) and stay time (between two \texttt{<|time|>} tokens) of your next activity.
    
    \vspace{0.4em}
    \hrule
    \vspace{0.4em}
    
    \textbf{\textsc{[User]}} \\
    \textbf{\#\#\# My Diaries:} \\
    \textit{(Past diaries retrieved via BM25 based on today's activity context)} \\
    \texttt{\{\{Retrieved\_Diary\_Entries\}\}}
    
    \vspace{0.3em}
    \textbf{\#\#\# Today's Activities:} \\
    \texttt{\{\{Today\_Movement\_History\}\}}
    
    \vspace{0.3em}
    Currently it is \texttt{\{\{Current\_Time\}\}}, I am planning to go to ...
    
    \vspace{0.4em}
    \hrule
    \vspace{0.4em}
    
    \textbf{\textsc{[Assistant]}} \\
    \texttt{<|place|>\{\{Next\_Location\}\}<|place|>}, and stay for \texttt{<|time|>\{\{Stay\_Duration\}\}<|time|>} hours.
\end{tcolorbox}
\end{center}

To ensure the fairness of the benchmark, identical prompts are provided to all evaluated human-like agents. It is worth noting that while the prompt includes value preference information, the Role-Play Baseline does not utilize this information during reasoning, whereas other baselines, including our CVAgent, explicitly leverage these value preferences for decision-making.

\subsection{Prompt-Driven Reasoning Agents}

Building upon the role-playing paradigm, numerous studies have sought to elicit internal deliberation within human-like agents through specialized prompting techniques. We categorize this approach as the \textit{prompt-driven value-reasoning} baseline, characterized by adjustable reasoning intensity. The underlying decision-making process typically follows a \textit{generate-then-evaluate} pipeline: the agent first proposes a set of candidate actions and subsequently selects the optimal one based on its projected impact on the agent's internal mental state. Generally, the reasoning strength modulates the search space; higher strength leads to a broader and deeper exploration of candidate actions. The algorithmic logic of this prompt-driven reasoning agent is detailed in the following pseudocode (see Algorithm~\ref{alg:reasoning_agent}).

In our experiments, we set the candidate count to $K = 3$. This configuration is empirically grounded to mimic the limited cognitive capacity of humans when weighing options during a single thought process. For the baseline representing "zero reasoning strength," we utilize the initial output $A$ derived from $A \leftarrow \text{rp}(C_w, C_l, V)$ in Algorithm \ref{alg:reasoning_agent}. Notably, unlike the vanilla role-play baseline, this variant is explicitly conditioned on value preference information. To investigate the effects of increased deliberation, we vary the reasoning strength $T$ across $\{0, 1, 2, 4, 8\}$, simulating different levels of cognitive thoroughness and the resulting diversity in the sampled action space.

\subsection{Training Required Agents}

We utilized the CVA Bench training dataset to establish a standard baseline for human-like agent training via a two-stage pipeline: Supervised Fine-Tuning (SFT) followed by Direct Preference Optimization (DPO). Both stages were trained for a single epoch.

\begin{itemize}
    \item \textbf{Supervised Fine-Tuning (SFT):} The model was fine-tuned using the role-playing system prompts as context $x$ and the ground truth human behaviors as the target response $y_{gt}$. We employed the standard Cross-Entropy (CE) loss for optimization.
    
    \item \textbf{DPO Preference Construction \& Training:} To construct the preference dataset $\mathcal{D}_{\text{DPO}} = \{(x, y_w, y_l)\}$, we sampled $K=10$ candidate actions $\mathcal{Y} = \{y_1, \dots, y_K\}$ from the SFT-tuned model $\pi_{\text{SFT}}(\cdot|x)$ with a temperature of $0.8$. We then evaluated the similarity between each candidate and the ground truth using linguistic metrics, denoted as $S(y_i, y_{gt})$. The positive (chosen) sample $y_w$ and negative (rejected) sample $y_l$ were selected as:
    \begin{equation}
        \begin{aligned}
            y_w &= \operatorname*{argmax}_{y_i \in \mathcal{Y}} S(y_i, y_{gt}) \\
            y_l &= \operatorname*{argmin}_{y_i \in \mathcal{Y}} S(y_i, y_{gt})
        \end{aligned}
    \end{equation}
    During the DPO training phase, we utilized a hybrid loss function combining the standard sigmoid DPO loss ($\mathcal{L}_{\text{DPO}}$), a BCO pair loss ($\mathcal{L}_{\text{BCO}}$), and an SFT regularization term ($\mathcal{L}_{\text{SFT}}$). The total objective is defined as:
    \[
        \mathcal{L}_{\text{total}} = 1.0 \cdot \mathcal{L}_{\text{DPO}} + 0.2 \cdot \mathcal{L}_{\text{BCO}} + 1.2 \cdot \mathcal{L}_{\text{SFT}}
    \]
    Specific details on the linguistic metrics are provided in Tables 3 and 6. For reproducibility, our implementation is available in the supplementary materials.
\end{itemize}

\section{Psychological Bias of Group Simulation in Prompt-Driven Humanoid Agents Experiment Details}
\label{appendix:exp3.1_details}

\begin{figure*}[htbp]
    \centering
    \includegraphics[width=1.0\linewidth]{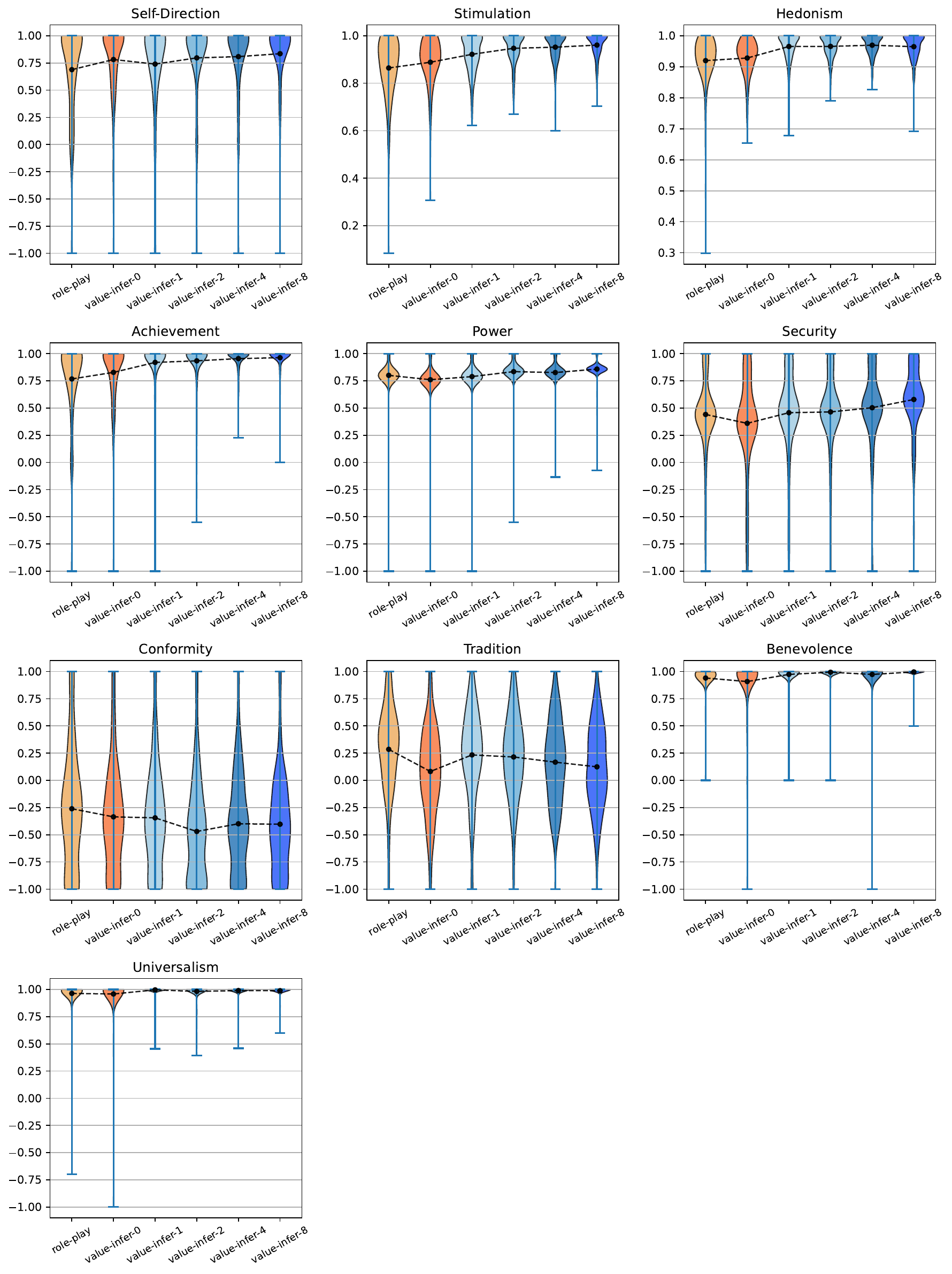}
    \caption{
        \textbf{Violin plots of population preference distributions across 10 Schwartz value dimensions.}
        This visualization presents the simulation results of the baseline methods (\texttt{role-play} and \texttt{value\_infer}) on CVABench, grouped by psychological measurement dimensions.
        The y-axis represents the normalized value preference score within the range $[-1, 1]$, and the black dashed line tracks the mean score across settings.
    }
    \label{fig:exp3.1 violin}
\end{figure*}

\definecolor{softred}{RGB}{250, 220, 220}  
\definecolor{softblue}{RGB}{225, 240, 255} 

\begin{table*}[htbp]
    \centering
    \caption{
    \textbf{Analysis of population Rigidity and Polarization under varying Value Inference strengths.} 
    This table presents a comprehensive analysis of how increasing inference strength (from \texttt{value\_infer\_0} to \texttt{8}) impacts the simulated population's value preference distribution.
    Due to the high dimensionality, the 10 value dimensions are split into two sub-tables: \textit{Part I} (Self-Direction to Power) and \textit{Part II} (Security to Universalism).
    \textbf{Panel A} reports the \textbf{Standard Deviation} (relative to Ground Truth). The monotonic decrease in variance indicates a trend towards \textbf{rigidity} and mode collapse, where the population loses its natural diversity.
    \textbf{Panel B} displays the \textbf{Absolute Mean Difference} relative to GT ($|\text{Model}| - |\text{GT}|$). Positive values indicate \textbf{polarization}, where the model's average preference becomes more extreme (closer to $+1$ or $-1$) than the human baseline.
    \textbf{Color Coding \& Conclusion:} In the \textbf{Average} column, \colorbox{softblue}{Blue} denotes insufficient guidance (Std $> 100\%$), while \colorbox{softred}{Red} highlights the dominant trend of overfitting.
    Crucially, the results demonstrate that higher inference strength is \textbf{detrimental}: it simultaneously causes the population to become \textbf{rigid} (vanishing variance in Panel A) and \textbf{polarized} (extreme mean deviation in Panel B), failing to preserve the nuance of the ground truth distribution.
    }
    \label{tab:exp3.1valueresult}
    
    \small 
    
    \textbf{\large Panel A: Variance Analysis (Standard Deviation relative to GT)} \par\medskip
    
    \textit{Part I: Self-Direction to Power} \par\smallskip
    \begin{tabular*}{\textwidth}{@{\extracolsep{\fill}}lcccccc}
        \toprule
        Model & Self-Direction & Stimulation & Hedonism & Achievement & Power & \textbf{Average} \\
        \midrule
        Role-Play Agent       & 109.32\% & 100.67\% & 75.10\%  & 250.61\% & 36.46\% & \cellcolor{softblue}110.29\% \\
        Reasoning Agent - 0 & 70.62\%  & 69.70\%  & 32.93\%  & 139.33\% & 40.65\% & \cellcolor{softblue}107.06\% \\
        Reasoning Agent - 1 & 107.26\% & 39.37\%  & 16.28\%  & 84.81\%  & 37.30\% & \cellcolor{softred}75.67\% \\
        Reasoning Agent - 2 & 85.90\%  & 26.58\%  & 13.95\%  & 65.02\%  & 25.62\% & \cellcolor{softred}64.69\% \\
        Reasoning Agent - 4 & 72.54\%  & 25.71\%  & 10.57\%  & 18.87\%  & 22.41\% & \cellcolor{softred}59.26\% \\
        Reasoning Agent - 8 & 69.78\%  & 18.89\%  & 15.27\%  & 23.56\%  & 9.04\%  & \cellcolor{softred}50.84\% \\
        \bottomrule
    \end{tabular*}

    \vspace{0.2cm}

    \textit{Part II: Security to Universalism} \par\smallskip
    \begin{tabular*}{\textwidth}{@{\extracolsep{\fill}}lcccccc}
        \toprule
        Model & Security & Conformity & Tradition & Benevolence & Universalism & \textbf{Average} \\
        \midrule
        Role-Play Agent       & 145.33\% & 185.22\% & 123.79\% & 25.27\%  & 51.15\%  & \cellcolor{softblue}110.29\% \\
        Reasoning Agent - 0 & 218.70\% & 151.78\% & 131.83\% & 87.45\%  & 127.63\% & \cellcolor{softblue}107.06\% \\
        Reasoning Agent - 1 & 172.52\% & 144.77\% & 135.92\% & 15.53\%  & 2.97\%   & \cellcolor{softred}75.67\% \\
        Reasoning Agent - 2 & 147.81\% & 143.44\% & 119.97\% & 4.71\%   & 13.91\%  & \cellcolor{softred}64.69\% \\
        Reasoning Agent - 4 & 153.55\% & 130.42\% & 114.53\% & 38.37\%  & 5.63\%   & \cellcolor{softred}59.26\% \\
        Reasoning Agent - 8 & 124.51\% & 126.14\% & 114.71\% & 1.30\%   & 5.22\%   & \cellcolor{softred}50.84\% \\
        \bottomrule
    \end{tabular*}

    \vspace{0.8cm} 

    \textbf{\large Panel B: Mean Analysis (Absolute Mean Difference relative to GT)} \par\medskip

    \textit{Part I: Self-Direction to Power} \par\smallskip
    \begin{tabular*}{\textwidth}{@{\extracolsep{\fill}}lcccccc}
        \toprule
        Model & Self-Direction & Stimulation & Hedonism & Achievement & Power & \textbf{Average} \\
        \midrule
        Role-Play Agent       & 0.0080 & -0.0269 & 0.0285 & -0.0850 & 0.0912 & \cellcolor{softred}0.0173 \\
        Reasoning Agent - 0 & 0.1013 & -0.0028 & 0.0369 & -0.0246 & 0.0511 & \cellcolor{softred}0.0071 \\
        Reasoning Agent - 1 & 0.0592 & 0.0306  & 0.0741 & 0.0672  & 0.0798 & \cellcolor{softred}0.0583 \\
        Reasoning Agent - 2 & 0.1163 & 0.0560  & 0.0744 & 0.0815  & 0.1254 & \cellcolor{softred}0.0844 \\
        Reasoning Agent - 4 & 0.1279 & 0.0607  & 0.0782 & 0.1020  & 0.1170 & \cellcolor{softred}0.0785 \\
        Reasoning Agent - 8 & 0.1547 & 0.0694  & 0.0734 & 0.1103  & 0.1486 & \cellcolor{softred}0.0914 \\
        \bottomrule
    \end{tabular*}

    \vspace{0.2cm}

    \textit{Part II: Security to Universalism} \par\smallskip
    \begin{tabular*}{\textwidth}{@{\extracolsep{\fill}}lcccccc}
        \toprule
        Model & Security & Conformity & Tradition & Benevolence & Universalism & \textbf{Average} \\
        \midrule
        Role-Play Agent       & -0.0815 & 0.0301 & 0.1139  & 0.0568 & 0.0381 & \cellcolor{softred}0.0173 \\
        Reasoning Agent - 0 & -0.1632 & 0.1047 & -0.0896 & 0.0243 & 0.0327 & \cellcolor{softred}0.0071 \\
        Reasoning Agent - 1 & -0.0653 & 0.1138 & 0.0627  & 0.0901 & 0.0709 & \cellcolor{softred}0.0583 \\
        Reasoning Agent - 2 & -0.0581 & 0.2388 & 0.0435  & 0.1100 & 0.0563 & \cellcolor{softred}0.0844 \\
        Reasoning Agent - 4 & -0.0195 & 0.1682 & -0.0042 & 0.0898 & 0.0646 & \cellcolor{softred}0.0785 \\
        Reasoning Agent - 8 & 0.0562  & 0.1723 & -0.0469 & 0.1129 & 0.0629 & \cellcolor{softred}0.0914 \\
        \bottomrule
    \end{tabular*}
\end{table*}

\begin{table*}[h]
    \centering
    \caption{Linguistic evaluation results across two simulation scenarios: \textit{Media Review} and \textit{Online Conversation}. The metrics report the Wasserstein Distance (WD) between the generated content and human ground truth. \textbf{{\color{red}Lower} values indicate better alignment with human linguistic patterns.}}
    \label{tab:linguistic_metrics}
    
    \renewcommand{\arraystretch}{1.2}
    \setlength{\tabcolsep}{9pt}
    
    \begin{tabular}{lccccccc}
        \toprule
        \multirow{2}{*}{\textbf{Model}} & \multicolumn{2}{c}{\textbf{Length (WD)}} & \multirow{2}{*}{\textbf{TTR (WD)}} & \multicolumn{4}{c}{\textbf{POS Tags (WD)}} \\
        \cmidrule(lr){2-3} \cmidrule(lr){5-8}
         & Doc-Len & Avg-Len &  & Adj & Adv & Noun & Verb \\
        \midrule
        
        \multicolumn{8}{c}{\textbf{\textit{Dataset I: Media Review Simulation}}} \\
        \midrule
        Role-Play Agent     & 273.95 & 3.89 & {\color{blue}0.06} & 6.46 & {\color{blue}2.45} & 8.75 & 5.69 \\
        Reasoning Agent - 0   & 340.68 & 4.30 & 0.06 & 7.65 & 2.63 & 11.03 & 7.07 \\
        Reasoning Agent - 2   & 250.04 & {\color{blue}2.66} & 0.06 & {\color{blue}3.40} & 3.75 & 8.57 & 5.86 \\
        Reasoning Agent - 4   & {\color{blue}239.31} & 3.29 & 0.06 & \textbf{3.28} & 3.43 & {\color{blue}8.07} & {\color{blue}5.67} \\
        \textbf{Ours} & \textbf{156.03} & \textbf{1.81} & \textbf{0.04} & 3.68 & \textbf{2.20} & \textbf{6.40} & \textbf{1.60} \\
        
        \midrule
        \multicolumn{8}{c}{\textbf{\textit{Dataset II: Online Conversation Simulation}}} \\
        \midrule
        Role-Play Agent     & 267.99 & {\color{blue}4.33} & 0.07 & 5.25 & 3.50 & 9.29 & 7.09 \\
        Reasoning Agent - 0   & 326.32 & 4.87 & 0.08 & 6.34 & 3.92 & 11.44 & 8.58 \\
        Reasoning Agent - 2   & {\color{blue}245.79} & 4.70 & {\color{blue}0.06} & {\color{blue}4.25} & 2.41 & {\color{blue}7.87} & {\color{blue}5.98} \\
        Reasoning Agent - 4   & 268.59 & 4.93 & 0.06 & 4.75 & {\color{blue}2.39} & 8.74 & 6.49 \\
        \textbf{Ours} & \textbf{106.98} & \textbf{1.83} & \textbf{0.04} & \textbf{1.47} & \textbf{0.97} & \textbf{3.62} & \textbf{2.23} \\
        
        \bottomrule
    \end{tabular}
\end{table*}

The baseline experiments conducted in CVABench quantify the underlying operational logic of current mainstream human-like agents. 
Our findings suggest that relying solely on an LLM’s instruction-following and in-context learning (ICL) capabilities for prompt-driven inner state modeling is inadequate; it lacks precision at the individual level and exhibits biases across group-level psychological indicators. 
While individual-level accuracy is detailed in the main text, this section provides a comprehensive presentation of group-level results. 
We include data for all ten value dimensions to supplement the single representative dimension shown in the main paper due to space constraints.

\paragraph{Observation: Increasing Reasoning Depth Leads to Extreme Value Polarization and Reduced Variance.}
We analyzed the iterative value reasoning process in prompt-driven agents (see Appendix~\ref{appendix:baselines}). As the number of reasoning iterations increased ($k \in \{0, 1, 2, 4, 8\}$), we observed two clear trends in the measured value-alignment scores:

\begin{itemize}
    \item \textbf{Reduced Value Variance:} The variance of the alignment scores decreased substantially as $k$ increased. This trend shows that the iterative reasoning procedure progressively narrows the range of the agent’s outputs. The model's responses become more predictable and lose their original diversity.
    
    \item \textbf{Increased Polarization:} At the same time, the alignment scores moved toward the extreme values of $-1$ and $1$. In this context, a score of $1$ represents strong support for a value (high consistency), while $-1$ represents strong opposition (high inconsistency). Instead of remaining neutral, the model's positions became increasingly binary.
\end{itemize}

These observations are quantitatively corroborated in \textbf{Table~\ref{tab:exp3.1valueresult}} and visually illustrated in \textbf{Figure~\ref{fig:exp3.1 violin}}. Specifically, Panel A of Table~\ref{tab:exp3.1valueresult} demonstrates a monotonic decrease in standard deviation across dimensions, confirming the collapse of population diversity. Concurrently, Panel B reveals that the model's average value preferences deviate significantly from the ground truth, shifting towards extreme bounds. The violin plots in Figure~\ref{fig:exp3.1 violin} further visualize this trajectory, showing how the distribution morphs from a diverse spread into narrow, rigid modes as reasoning strength increases.

This phenomenon highlights a critical failure of prompt-driven value reasoning. Real human behavior is complex and often reflects conflicting values. However, repeated and explicit reasoning pushes the LLM toward a polarized state. When the model is forced to reason more deeply about its values, it defaults to simplified and exaggerated positions. Consequently, it fails to capture the subtle and moderate value alignment that characterizes real human behavior.

\begin{figure*}[h]
    \centering
    \includegraphics[width=1.0\linewidth]{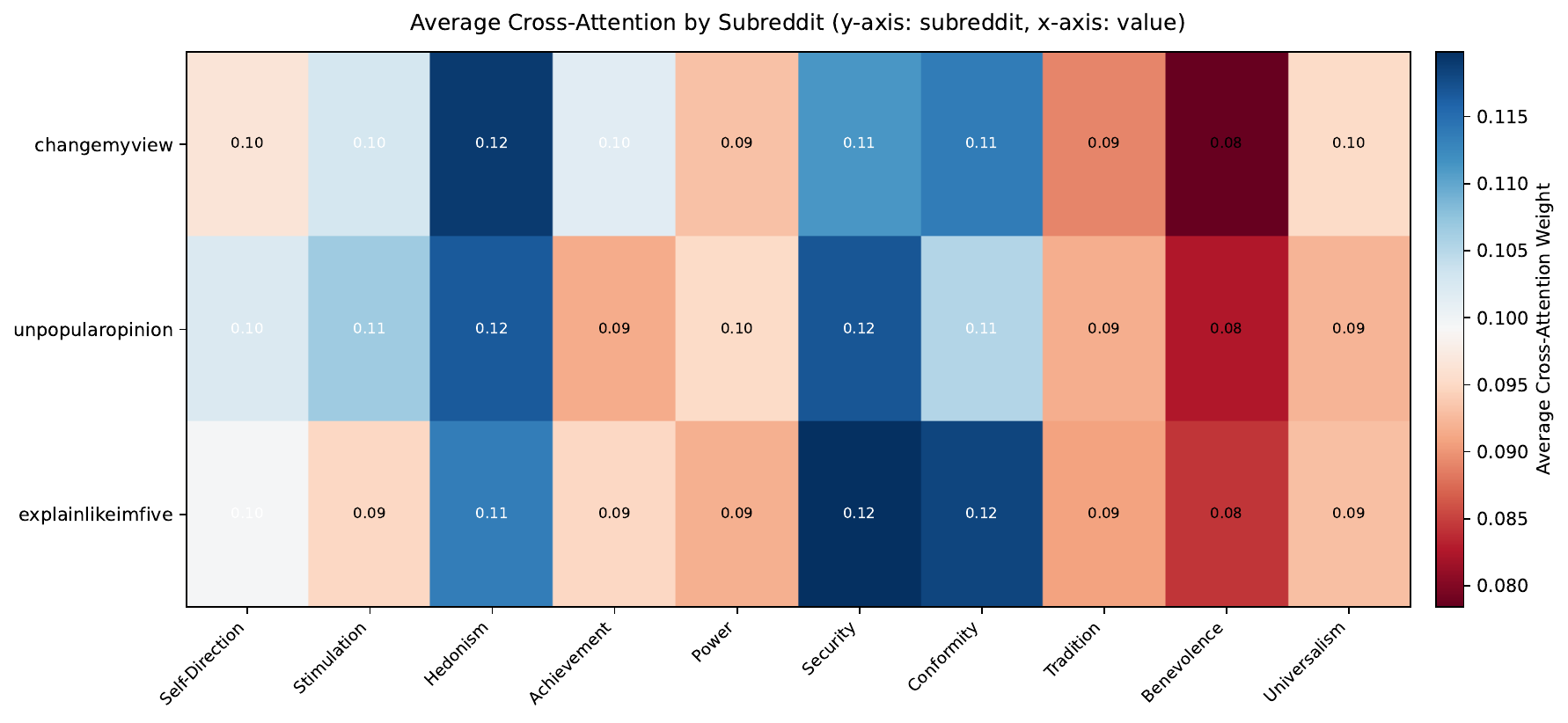}
    \caption {\textbf{Coarse-grained value activation analysis by subreddit.} This heatmap displays the average cross-attention weights assigned to each of the ten Schwartz values across three distinct Reddit communities (\textit{changemyview}, \textit{unpopularopinion}, \textit{explainlikeimfive}). The color intensity reflects the magnitude of attention, where blue indicates higher activation (e.g., \textit{Hedonism}, \textit{Security}) and red indicates lower activation (e.g., \textit{Benevolence}). The relatively consistent activation patterns across diverse subreddits suggest that community-level metadata is insufficient to fully disentangle value triggers, motivating the need for the fine-grained, token-level projection analysis presented in the subsequent figures.}
    \label{fig:exp3.3subreddit-analyze}
\end{figure*}

\begin{figure*}[htbp]
    \centering
    \includegraphics[width=1.0\linewidth]{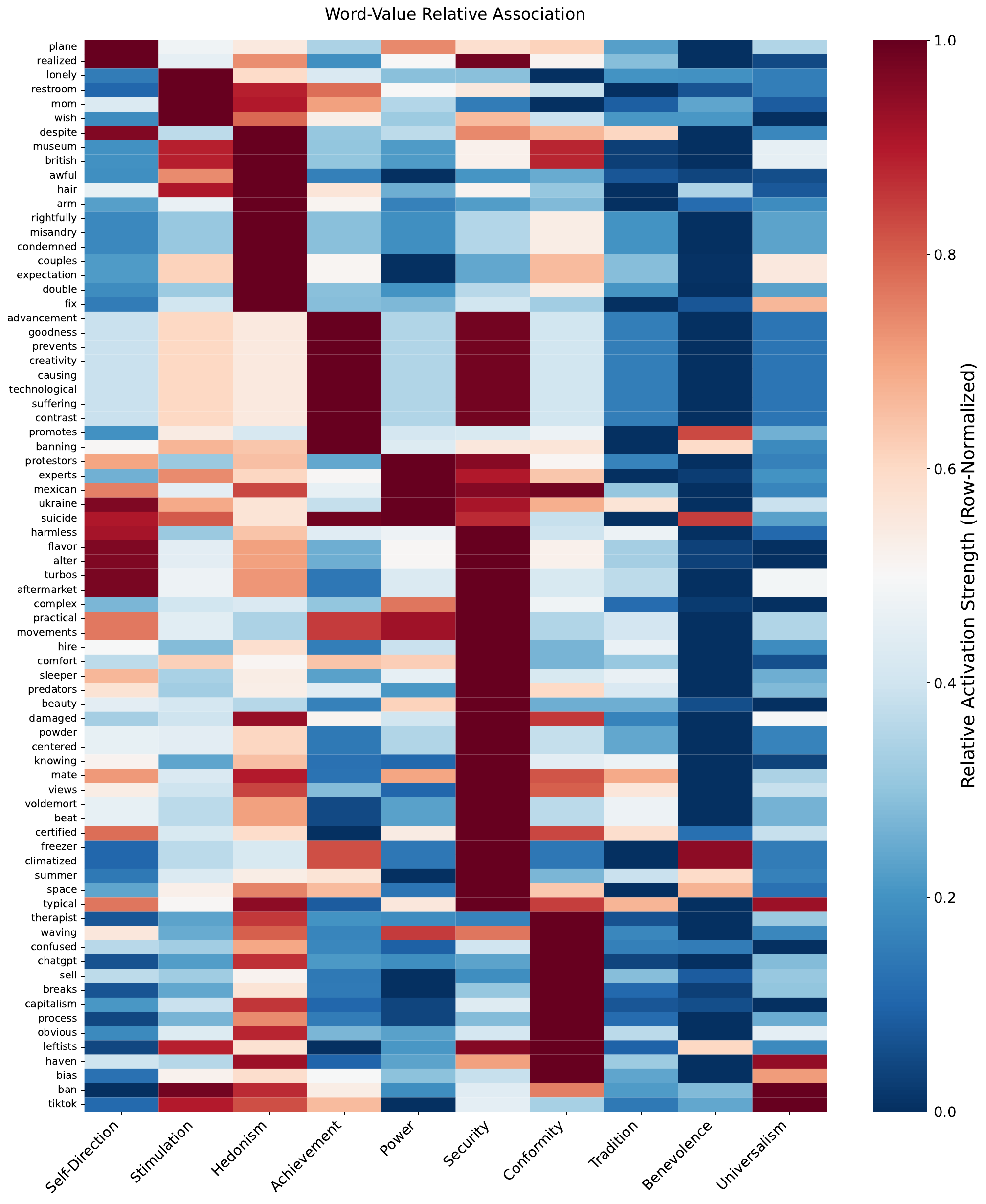}
    \caption {This matrix illustrates the \textit{relative affinity} between discriminative words (y-axis) and value dimensions (x-axis). Words are sorted by their primary value association. The color scale represents the row-normalized score $\hat{S}(w, v)$, where dark red (1.0) indicates the value dimension that a specific word activates most strongly relative to others. The clear diagonal structure confirms that the model learns specific, non-overlapping lexical mappings for different values.}
    \label{fig:exp3.3heatmap}
\end{figure*}

\begin{figure*}[h]
    \centering
    \includegraphics[width=1.0\linewidth]{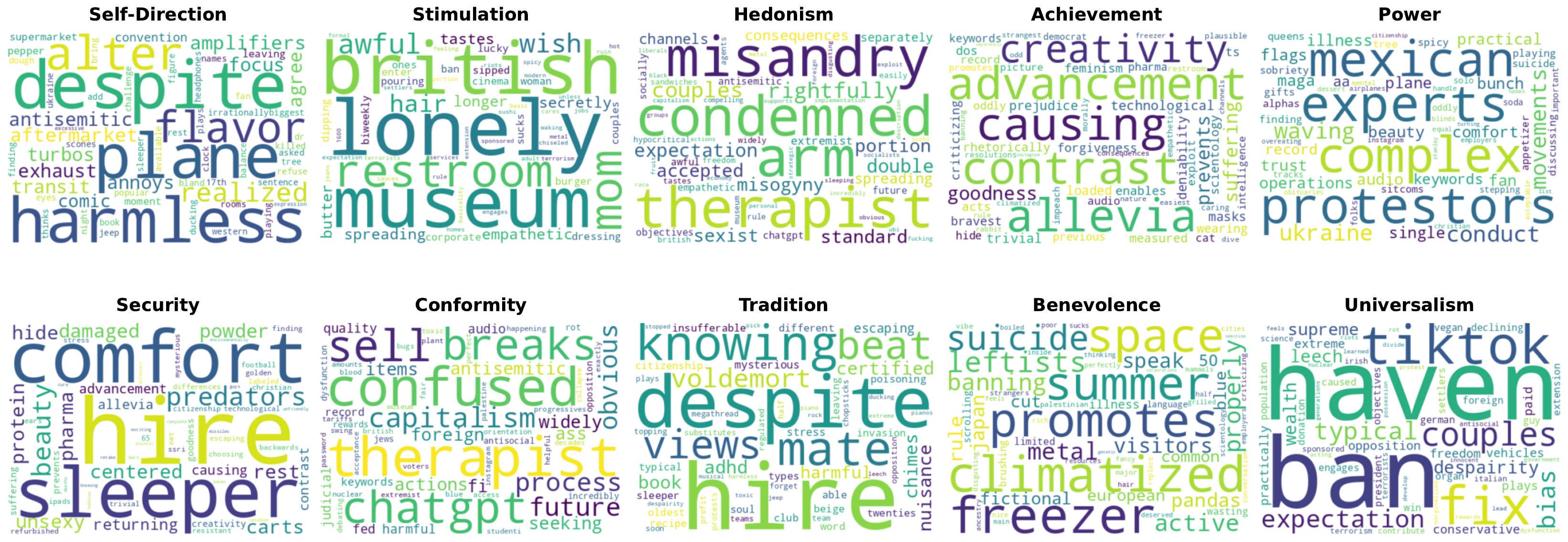}
    \caption {Each subplot visualizes the most representative vocabulary for a specific value dimension. Word sizes reflect the aggregated attention magnitude derived from the validation set. This comparison highlights distinct semantic clusters—such as "creativity" for \textit{Achievement} versus "confused" for \textit{Conformity}—demonstrating the model's capability to capture value-specific context.}
    \label{fig:exp3.3wordclouds-all}
\end{figure*}

\section{CVABench Metrics and Detailed Results}
\label{appendix:exp3.2}

\subsection{Metrics Explanation}

Let $N$ denote the total number of samples in the simulation set of CVA Bench.

\paragraph{Social Media Review}
In the social media domain, the model predicts a user's discrete rating $r \in \{1, \dots, 5\}$ and review sentiment $s$ based on historical context. 
We evaluate the predictive performance using standard Accuracy:
\begin{equation}
\text{Acc} = \frac{1}{N} \sum_{i=1}^{N} \mathbb{I}(\hat{y}_i = y_i)
\end{equation}
where $\hat{y}_i$ and $y_i$ represent the predicted and ground-truth labels (rating or sentiment), and $\mathbb{I}(\cdot)$ denotes the indicator function.

Beyond semantic accuracy, we assess the \textit{linguistic fidelity} by comparing the lexical richness of generated reviews against real user data. 
We calculate the Type-Token Ratio (TTR) for the generated and ground-truth corpora, respectively. 
The discrepancy between the two TTR distributions, denoted as $P_{\text{gen}}$ and $P_{\text{real}}$, is quantified using the 1-Wasserstein Distance:
\begin{equation}
W_1(P_{\text{gen}}, P_{\text{real}}) = \inf_{\gamma \in \Gamma(P_{\text{gen}}, P_{\text{real}})} \mathbb{E}_{(x, y) \sim \gamma} [|x - y|]
\end{equation}
where $\Gamma(P_{\text{gen}}, P_{\text{real}})$ is the set of all joint distributions $\gamma(x, y)$ whose marginals are $P_{\text{gen}}$ and $P_{\text{real}}$, and $x, y$ represent the TTR values from the simulated and real distributions, respectively.

\paragraph{Conversation Discourse}
In this domain, we evaluate the agent's capacity to replicate specific user stances across different subreddits. 
The primary metric, Attitude Accuracy ($Acc_{att}$), is formulated analogously to Eq.~(1), where $y_i$ denotes the discrete attitude label extracted from the user's ground-truth comments. 
Mirroring the social media evaluation, we further assess linguistic fidelity by quantifying the discrepancy between the TTR distributions of generated and authentic comments using the 1-Wasserstein Distance.

\paragraph{Spatio-Temporal Mobility}
This task involves predicting the next Point-of-Interest (POI) category $c$ and the duration of stay $t$. We utilize Category Accuracy ($Acc_{cat}$) for spatial prediction and Mean Squared Error (MSE) for temporal precision:
\begin{equation}
\text{MSE} = \frac{1}{N} \sum_{i=1}^{N} (\hat{t}_i - t_i)^2
\end{equation}
where $\hat{t}_i$ and $t_i$ denote the predicted and actual stay durations (in minutes), respectively.

\paragraph{Value Distribution Variance (Var\%)}
To quantitatively assess the alignment of the simulated population's diversity with real-world ground truth, we introduce the Value Distribution Variance metric. This metric measures the relative deviation between the variance of the value alignment scores generated by the agents and that of the empirical human data across the three domains. Formally, let $\sigma^2_{sim}$ denote the variance of the simulated value distribution and $\sigma^2_{gt}$ denote the variance of the ground truth distribution. The metric is calculated as the percentage difference:
\begin{equation}
    \text{Var\%} = \frac{\sigma^2_{sim} - \sigma^2_{gt}}{\sigma^2_{gt}} \times 100\%
\end{equation}
The interpretation of this metric is as follows:
\begin{itemize}
    \item \textbf{Positive Values (+):} A positive deviation ($\sigma^2_{sim} > \sigma^2_{gt}$) indicates that the simulated population exhibits excessive variance compared to reality. This suggests ineffective modeling, where agents produce unstable or overly random behaviors that fail to capture consistent human traits.
    \item \textbf{Negative Values (-):} A negative deviation ($\sigma^2_{sim} < \sigma^2_{gt}$) indicates a collapse in variance. This corresponds to the \textit{value polarization} or \textit{behavioral rigidity} phenomenon discussed in the main text, where agents converge to a narrow, stereotypical range of values and fail to reflect the richness of human diversity.
    \item \textbf{Optimal Alignment:} A value closer to $0$ represents superior alignment, indicating that the architecture successfully reproduces the natural degree of value diversity inherent in the real-world population.
\end{itemize}

\subsection{Detailed Results}

CVABench provides a comprehensive suite of linguistic metrics designed to analyze social media dynamics and online conversational patterns. Table~\ref{tab:linguistic_metrics} extends the analysis presented in Table~\ref{tab:main_exp_result} of the main text by incorporating additional linguistic indicators. All metrics detailed in Table~\ref{tab:linguistic_metrics} are quantified using the Wasserstein Distance (WD), defined as follows:

\begin{equation}
    W(P, Q) = \inf_{\gamma \in \Gamma(P, Q)} \mathbb{E}_{(x, y) \sim \gamma} [\|x - y\|]
    \label{eq:wasserstein}
\end{equation}

\noindent where $P$ and $Q$ represent the two probability distributions being compared, and $\Gamma(P, Q)$ denotes the set of all joint distributions $\gamma(x, y)$ whose marginals are $P$ and $Q$, respectively.

We evaluated a total of seven metrics. The first two focus on overall sentence structure: specifically, the WD between the distribution of review lengths and the distribution of average sentence lengths. The remaining five metrics assess lexical composition, including lexical richness (measured by Type-Token Ratio, TTR) and the word frequency distributions of nouns, adverbs, adjectives, and verbs.

Consistent with the findings in the main text, we observe a similar phenomenon: CVA agents frequently outperform existing prompt-driven human-like agents, demonstrating smaller WD values relative to the ground-truth distribution. Counter-intuitively, however, increasing the number of reasoning rounds in prompt-driven agents does not necessarily yield performance closer to authentic human behavior.

\section{CVABench Case Study}
\label{appendix:exp3.3}

\subsection{Word-to-Value Projection Analysis}

To interpret which lexical features trigger specific value activations, we employ a co-occurrence-based projection method that maps the vocabulary space to the value space. Let $\mathcal{D} = \{x^{(1)}, x^{(2)}, \dots, x^{(N)}\}$ denote the corpus of $N$ context samples. For each input context $x^{(i)}$, the model produces a value activation vector $\mathbf{a}^{(i)} = [a_0^{(i)}, a_1^{(i)}, \dots, a_9^{(i)}]$, where $a_k^{(i)} \in [0, 1]$ represents the cross-attention weight (or activation intensity) for the $k$-th Schwartz value (e.g., $a_0$ for Self-Direction, $a_9$ for Universalism).

To quantify the contribution of a specific word $w$ to a value $v_k$, we first compute the Term Frequency-Inverse Document Frequency (TF-IDF) score, denoted as $t_{w, i}$, for word $w$ in context $x^{(i)}$. The TF-IDF weight serves to highlight distinctive terms while suppressing common stop words. The \textit{Word-Value Relevance Score}, $S(w, v_k)$, is calculated as the weighted average of the value activations across all contexts where the word appears, weighted by the word's textual importance:

\begin{equation}
    S(w, v_k) = \frac{\sum_{i=1}^{N} t_{w, i} \cdot a_k^{(i)}}{\sum_{i=1}^{N} t_{w, i} + \epsilon}
\end{equation}

where $\epsilon$ is a small smoothing term to avoid division by zero. 

Intuitively, $S(w, v_k)$ represents the expected activation level of value $v_k$ given the presence of word $w$. A high score indicates that whenever word $w$ appears in the context, the model consistently assigns a high attention weight to value $v_k$. To compare the relative semantic affinity of a word across different values, we further apply row-wise min-max normalization for visualization:

\begin{equation}
    \hat{S}(w, v_k) = \frac{S(w, v_k) - \min_{j} S(w, v_j)}{\max_{j} S(w, v_j) - \min_{j} S(w, v_j)}
\end{equation}

This normalization highlights the distinctive value preference of each word, allowing us to identify lexical markers specific to each value dimension.

\noindent \textbf{Visualization Strategy.} To interpret the learned associations intuitively, we employ two complementary visualization techniques:

\begin{enumerate}
    \item \textbf{Relative Activation Heatmap:} We construct a heatmap to analyze the discriminative power of lexical features across the ten value dimensions. In this visualization, the color intensity of a word $w$ corresponding to value $v_k$ is determined by the \textit{row-normalized} score $\hat{S}(w, v_k)$. By normalizing across the word dimension (row-wise), we filter out the effect of global word frequency and absolute model attention, focusing instead on the \textit{relative preference} of each word. A value of 1.0 (dark red) indicates the value dimension that the word $w$ most strongly activates relative to others, revealing the specific semantic alignment of the term.
    
    \item \textbf{Value-Specific Word Clouds:} To capture the overall semantic landscape of each value, we generate ten separate word clouds. Unlike the heatmap, the size of word $w$ in the cloud for value $v_k$ is proportional to its \textit{unnormalized} relevance score $S(w, v_k)$. This ensures that words with higher absolute activation weights—representing the core concepts that most strongly trigger the model's recognition of a specific value—are visually prominent.
\end{enumerate}

\begin{table*}[htbp]
    \centering
    \caption{Definitions of the Ten Basic Values in the Schwartz Value Theory}
    \label{tab:schwartz_values}
    \begin{tabularx}{\textwidth}{@{}lX@{}}
        \toprule
        \textbf{Value Dimension} & \textbf{Conceptual Definition} \\
        \midrule
        \textbf{Self-Direction} & Independent thought and action; choosing, creating, exploring. \\
        \textbf{Stimulation}    & Excitement, novelty, and challenge in life. \\
        \textbf{Hedonism}       & Pleasure and sensuous gratification for oneself. \\
        \textbf{Achievement}    & Personal success through demonstrating competence according to social standards. \\
        \textbf{Power}          & Social status and prestige, control or dominance over people and resources. \\
        \textbf{Security}       & Safety, harmony, and stability of society, of relationships, and of self. \\
        \textbf{Conformity}     & Restraint of actions, inclinations, and impulses likely to upset or harm others and violate social expectations or norms. \\
        \textbf{Tradition}      & Respect, commitment, and acceptance of the customs and ideas that traditional culture or religion provide the self. \\
        \textbf{Benevolence}    & Preservation and enhancement of the welfare of people with whom one is in frequent personal contact. \\
        \textbf{Universalism}   & Understanding, appreciation, tolerance, and protection for the welfare of all people and of nature. \\
        \bottomrule
    \end{tabularx}
\end{table*}

\subsection{Quantitative Analysis of the Circular Embedding Structure}

Our analysis reveals that the value embeddings learned by the \textbf{Value-Guided Verifier} exhibit a geometric topology that aligns remarkably well with the theoretical circular structure of the Schwartz Value System (see Table~\ref{tab:schwartz_values} for definitions). To rigorously quantify this phenomenon, we conducted a comparative analysis of the embedding topology before and after the verifier training phase.

\paragraph{Metric: Circular Inversion Score.} 
Since the Schwartz model is defined by the relative angular positions of values rather than a fixed linear order, we propose the \textbf{Circular Inversion Score (CIS)} as an evaluation metric.
First, for each value dimension $v_i$, we project its high-dimensional embedding $\mathbf{e}_v$ onto a 2D plane using Principal Component Analysis (PCA) and calculate its angular position $\theta_i$. This yields an observed circular sequence $\mathcal{S}_{obs}$ based on the sorted angles.
Let $\mathcal{S}_{gt}$ denote the theoretical order of the Schwartz circumplex. We quantify the structural fidelity using the \textbf{Circular Inversion Score (CIS)}, which measures the proportion of preserved pairwise relationships. First, we determine the minimum rotational inversion distance, denoted as $\mathcal{D}_{circ}$, by finding the optimal alignment that minimizes the number of pairwise swaps (Kendall's $\tau$ distance) between the observed sequence $\mathcal{S}_{obs}$ and the ground truth:

\begin{equation}
\begin{split}
    \mathcal{D}_{circ}&(\mathcal{S}_{obs}, \mathcal{S}_{gt}) \\
    &= \min_{k \in \{0, \dots, N-1\}} \mathcal{I} \left( \text{Rotate}_k(\mathcal{S}_{obs}), \mathcal{S}_{gt} \right)
\end{split}
\end{equation}

where $N$ is the number of values (e.g., $N=10$ for the standard SVS, or $N=8$ when excluding specific dimensions), $\text{Rotate}_k(\cdot)$ represents shifting the sequence by $k$ steps, and $\mathcal{I}(\cdot, \cdot)$ calculates the standard inversion count. The final CIS is then normalized as:
\begin{equation}
    \text{CIS} = 1 - \frac{\mathcal{D}_{circ}}{N(N-1)/2}
\end{equation}
representing the structural alignment accuracy, where 1.0 indicates a perfect reconstruction of the theoretical circular order.

\begin{table}[htbp]
    \centering
    \caption{Comparison of Circumplex Index of Structure (CIS) scores across different verification stages. The CIS metric evaluates the alignment with the theoretical circular value structure, where 1.00 represents a perfect fit.}
    \label{tab:cis_comparison}
    \begin{tabular}{lcc}
        \toprule
        \textbf{Method} & \textbf{CIS Score} \\
        \midrule
        Ground Truth & 1.00 \\
        Trained Verifier & 0.75 \\
        Original Verifier & 0.48 \\
        \bottomrule
    \end{tabular}
\end{table}

\paragraph{Results.} 
The trained verifier demonstrated a significant capability to recover the latent circular structure of human values, achieving a Circumplex Index of Structure (CIS) of 0.75. It is important to note that the dimensions of Power and Security were excluded from this topological analysis. These two dimensions were filtered out during the validity check phase, as their measurement outputs exhibited a noun probability exceeding 10\%, indicating unstable representation in the generated data compared to the stable adjectival descriptions of other values.

Structural Strengths: The relatively high CIS score suggests that the model successfully captured the coarse-grained relationships between value clusters. Specifically, the adjacency within the Self-Transcendence sector (Universalism and Benevolence) and the Conservation sector (Tradition and Conformity) was largely preserved. This indicates that the verifier has learned to distinguish between altruistic, social-focus values and self-restrictive, stability-focus values.

Deviations and Causes: The primary penalty to the CIS score arises from specific semantic entanglements. The most significant deviation is the displacement of Achievement, which migrated from the Self-Enhancement quadrant to a position adjacent to Self-Direction. This suggests that the model may conflate the semantic features of "individual success" (Achievement) with "intellectual autonomy" (Self-Direction). Additionally, the observed adjacency of Hedonism and Tradition implies that the current model struggles to fully disentangle the tension between gratification and restraint, possibly due to data sparsity in these specific opposing pairings.

\section{CVA Architecture Training Details}
\label{appendix:training_details}

\subsection{Generator Training: Value-Conditional De-biasing}
The training pipeline for the CVA generator aligns with the baseline algorithm but incorporates explicit \textit{value conditioning} to mitigate the inherent psychological biases of the base LLM.

\begin{itemize}
    \item \textbf{SFT Phase:} While the baseline attempts to approximate the standard context-action mapping $P_{\theta}(A|C)$, the CVA architecture learns a value-conditioned mapping $P_{\theta}(A|C, V)$. By injecting value labels $V$ alongside the context $C$, we guide the model to distinguish between generic responses and value-aligned behaviors.
    \item \textbf{DPO Phase:} Similarly, in the DPO stage, we utilize $(C, V)$ as joint inputs. This further refines the policy to align with specific value profiles, effectively decoupling the agent's generation process from the base model's prior biases.
\end{itemize}

To address computational constraints during the SFT and DPO phases, we employed memory-efficient techniques, including QLoRA and Flash Attention \citep{qlora, flashattn2}. Comprehensive details regarding training hyperparameters are available in the code repository.

\subsection{Verifier Architecture and Training}
To explicitly model human preference selection, we introduce a \textbf{Value-Guided Verifier}.

\paragraph{Architecture.} The verifier employs a multi-encoder design. A text encoder processes the action $A$ and context $C$ to obtain embeddings $E_a$ and $E_c$, respectively. Simultaneously, a value encoder maps the value profile $V$ to an embedding $E_v$. To capture the context-dependent nature of values, we employ a cross-attention mechanism where the context embedding guides the attention over the value embedding, producing a refined value representation $E'_v$:
\begin{equation}
    E'_v = \operatorname{CrossAttn}(Q=E_c, K=E_v, V=E_v)
\end{equation}
Finally, the concatenated vector $[E'_v; E_a]$ is fed into a Multi-Layer Perceptron (MLP) to predict a value consistency score $s(A, C, V)$.

\paragraph{Training Objective.} We construct the training dataset by sampling 5 candidate actions from the DPO-tuned model (at temperature 0.8) for each context. These candidates are ranked based on their linguistic similarity to the ground truth behavior. We denote a preferred action as $A_w$ and a rejected action as $A_l$. The selection process is defined as:
\begin{equation}
\small 
\begin{split}
    A_w &= \operatorname*{argmax}_{A_i \in \mathcal{Y}} S(A_i, A_{gt}) \\
    A_l &= \operatorname*{argmin}_{A_i \in \mathcal{Y}} S(A_i, A_{gt})
\end{split}
\end{equation}
The verifier is trained to maximize the margin between the scores of the preferred and rejected actions. We minimize the pairwise ranking loss:
\begin{equation}
\small 
\begin{split}
    \mathcal{L}_{\text{ver}} = - \mathbb{E}_{\mathcal{D}} \Big[ \log \sigma \big( &s(A_w, C, V) \\
    &- s(A_l, C, V) \big) \Big]
\end{split}
\end{equation}
where $\mathcal{D}$ represents the dataset tuples $(C, V, A_w, A_l)$ and $\sigma(\cdot)$ is the sigmoid function. This objective ensures that the verifier assigns strictly higher consistency scores to actions that better align with the human ground truth under the given value profile.

\end{document}